\documentclass[sigconf]{acmart}

\chardef\usethreeparttable=0
\chardef\usemathcommands=1
\chardef\usealgorithmicx=0
\chardef\usexcolor=0

\newcommand{\importifndef}[2]{\ifcsname#1\endcsname\else\usepackage{#2}\fi}

\if\the\usexcolor
\usepackage[dvipsnames, svgnames, x11names, table]{xcolor}
\fi

\importifndef{algorithm}{algorithm}
\usepackage[noend]{algpseudocode}

\if\the\usealgorithmicx1
\usepackage{algorithmicx}
\fi

\usepackage{amsmath,amsfonts,amsthm,bm}
\usepackage{nicefrac}

\usepackage{microtype}
\usepackage{wrapfig}
\usepackage{graphicx}
\usepackage{subcaption}

\usepackage{array}
\usepackage{diagbox}
\usepackage{multirow}
\usepackage{booktabs}

\if\the\usethreeparttable1
\usepackage{threeparttable}
\fi

\usepackage{tikz}
\usepackage{pgfplots}
\usetikzlibrary{plotmarks}
\pgfplotsset{compat=1.16}

\usepackage{listings}

\usepackage{hyperref}
\usepackage{enumitem}

\newcommand{\fixed}[1]{\iffalse fixed: #1\fi}
\newcommand{\rej}[1]{\iffalse#1\fi}

\if\the\usemathcommands1

\def\eqref#1{Eq.~(\ref{#1})}

\def\1{\bm{1}}

\DeclareMathAlphabet{\mathsfit}{\encodingdefault}{\sfdefault}{m}{sl}
\SetMathAlphabet{\mathsfit}{bold}{\encodingdefault}{\sfdefault}{bx}{n}

\newcommand{\parfrac}[2]{\frac{\partial{#1}}{\partial{#2}}}
\newcommand{\chainrule}[3]{\parfrac{#1}{#2}\cdot\parfrac{#2}{#3}}

\DeclareMathOperator*{\argmin}{arg\,min}

\fi

\AtBeginDocument{%
  \providecommand\BibTeX{{%
    \normalfont B\kern-0.5em{\scshape i\kern-0.25em b}\kern-0.8em\TeX}}}

\acmConference[The Fifth International Conference on Distributed Artificial Intelligence]{}{Nov 30th -- Dec 3rd, 2023}{Singapore}

\copyrightyear{2023}
\acmYear{2023}
\setcopyright{acmlicensed}\acmConference[DAI '23]{The Fifth International Conference on Distributed Artificial Intelligence}{November 30-December 3, 2023}{Singapore, Singapore}
\acmBooktitle{The Fifth International Conference on Distributed Artificial Intelligence (DAI '23), November 30-December 3, 2023, Singapore, Singapore}
\acmPrice{15.00}
\acmDOI{10.1145/3627676.3627684}
\acmISBN{979-8-4007-0848-0/23/11}

\begin{document}

\title{D4W: Dependable Data-Driven Dynamics for Wheeled Robots}

\author{Yunfeng Lin}
\orcid{0009-0004-8953-9244}
\email{linyunfeng@sjtu.edu.cn}
\affiliation{%
  \institution{Shanghai Jiao Tong University}
  \city{Minhang}
  \state{Shanghai}
  \country{China}
  \postcode{200240}
}

\author{Minghuan Liu}
\email{minghuanliu@sjtu.edu.cn}
\affiliation{%
  \institution{Shanghai Jiao Tong University}
  \city{Minhang}
  \state{Shanghai}
  \country{China}
  \postcode{200240}
}

\author{Yong Yu}
\email{yyu@apex.sjtu.edu.cn}
\affiliation{%
  \institution{Shanghai Jiao Tong University}
  \city{Minhang}
  \state{Shanghai}
  \country{China}
  \postcode{200240}
}

\renewcommand{\shortauthors}{Lin, et al.}

\begin{abstract}
Wheeled robots have gained significant attention due to their wide range of applications in manufacturing, logistics, and service industries.
However, due to the difficulty of building a highly accurate dynamics model for wheeled robots, developing and testing control algorithms for them remains challenging and time-consuming, requiring extensive physical experimentation.
To address this problem, we propose D4W, i.e., Dependable Data-Driven Dynamics for Wheeled Robots, a simulation framework incorporating data-driven methods to accelerate the development and evaluation of algorithms for wheeled robots.
The key contribution of D4W is a solution that utilizes real-world sensor data to learn accurate models of robot dynamics.
The learned dynamics can capture complex robot behaviors and interactions with the environment throughout simulations, surpassing the limitations of analytical methods, which only work in simplified scenarios.
Experimental results show that D4W achieves the best simulation accuracy compared to traditional approaches, allowing for rapid iteration of wheel robot algorithms with less or no need for fine-tuning in reality.
We further verify the usability and practicality of the proposed framework through integration with existing simulators and controllers.
\end{abstract}


\maketitle

\section{Introduction}

Wheeled mobile robots (WMR) play a crucial role in various domains, from industrial and agricultural automation to public services, due to their versatility and flexibility in dynamic environments~\cite{robothandbook}.
Designed to navigate independently around the work facilities, wheeled robots utilize proprioceptive sensory data and baseline maps to perform path planning and collision avoidance.
Although such robots have a simpler configuration space than limbed robots, developing effective control and navigation algorithms for them is still challenging, requiring accurate modeling of robot dynamics.

Traditional approaches often rely on analytical models, where a physical simulator computes a robot's trajectory under given commands in a virtual environment~\cite{mujoco, pybullet, isaacgym}.
The simulator typically has access to known properties of the robot, such as its mass, inertia, and geometry from the design schematics, predefined controller models, and parameters~\cite{diffcontroller}, which can be manually adjusted to match reality.
While analytical methods are theoretically accurate, they may not capture the intricate dynamics of real-world systems, such as skids and slides.
As a result, the performance of algorithms evaluated on these models may be sub-optimal in practice, leading to reduced efficiency and increased safety risks.
Eventually, manual alignment with real-world dynamics is required to guarantee the robot's usability, which becomes a bottleneck in the algorithm development.

A sufficiently accurate dynamics model of wheeled robots is required to address the challenges above. To this end, we propose a framework named D4W (Dependable Data-Driven Dynamics for Wheeled Robots), designed to extract the underlying complex non-linear relationships from real-world observations by combining physics-based simulation with data-driven methods.
Specifically, D4W automates the data-gathering procedure that builds the dataset in an efficient and unattended manner. It makes the robot record its states while carrying out a standard sequence, allowing it to traverse the reachable areas without triggering collisions.
To improve the explainability and generalizability of the learned dynamics, we perform egocentric transformations on model inputs in each simulation step of D4W.
This guarantees translational and rotational symmetry in space and translational symmetry in time.
Furthermore, D4W provides interoperability with existing simulator and controller implementations, enabling a seamless transition to data-driven dynamics while retaining well-defined functionalities such as rendering~\cite{gradsim, orbit}.




In summary, the contributions of this paper lie in the following:
\begin{itemize}[leftmargin=15pt]
    \item We propose D4W, a generalized framework for learning an accurate dynamics model by minimizing the difference between simulated trajectories and the observed ones while satisfying necessary kinematic invariants, which sets up a routine to gather real-world robot motion and sensory data automatically and efficiently.
    \item To our knowledge, this is the first work realizing an interoperable dynamics simulator augmented with neural networks and trained on real-world observations in the field of wheeled mobile robots.
    \item The parameterized dynamics model trained by D4W achieves the best simulation accuracy compared with existing analytical simulators.
\end{itemize}

\section{Preliminaries}


\begin{table}[htb]
    \centering
    \caption{Notations for the WMR dynamics formulation.}
    \label{tab:_nomenclature}
    \small
    \begin{tabular}{lp{5cm}}
        \toprule
        Notation & Description\\
        \midrule
        $P$ & the coordinate reference point of the robot \\
        $x$ & x-axis coordinate of $P$ in a reference frame \\
        $y$ & y-axis coordinate of $P$ in a reference frame \\
        $\theta$ & orientation of the robot \\
        $s$ & robot chassis speed in longitudinal direction \\
        $\omega$ & instantaneous angular velocity of the chassis \\
        $r$ & radius of a driving wheel \\
        $R$ & distance between $P$ to each driving wheel \\
        $w_l, w_r$ & angular velocity of the left and right wheels \\
        $s^c$ & robot speed command \\
        $\omega^c$ & robot rotational speed command \\
        \bottomrule
    \end{tabular}
\end{table}

We summarize the notations used in the formulations in ~Table~\ref{tab:_nomenclature}.

\begin{figure}[t]
\centering
\includegraphics[width=.8\columnwidth]{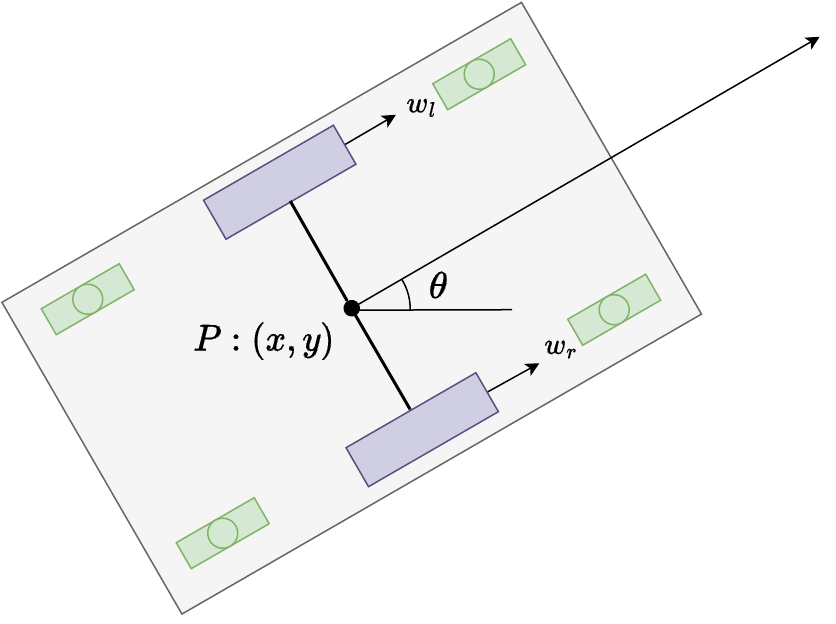}
\caption{
Diagram of a unicycle-type wheeled robot.
It shows the coordinate reference point $P$, the robot orientation $\theta$, and wheel speeds $w_l$ and $w_r$.
}
\label{fig:_wmr}
\end{figure}

\paragraph{Wheeled mobile robot.}
As depicted in~Figure~\ref{fig:_wmr}, a conventional wheeled mobile robot (WMR) comprises a rigid chassis and several non-deformable wheels rotating vertically around their axles.
The orientations of the wheels can be unconstrained or fixed, corresponding to different categories of wheeled robots: car-like robots with steering wheels and unicycle-like ones with fixed wheels on a shared axle.
Some wheels are actuated with active drive, while passive wheels are added for balance and stability.

This work focuses on unicycle-type WMRs moving on a flat (Euclidean) horizontal plane.
Specifically, the robot models used for evaluations include two active wheels fixed in symmetrical side positions and four passive caster wheels in every corner.

The state (posture) of a robot's body can be defined in an inertial frame as a three-dimensional vector:
\begin{equation}
    \mathbf{q} = (x, y, \theta)^T~,
\end{equation}
where $x$ and $y$ are the robot coordinates from the origin, and $\theta$ represents the orientation of its chassis.
The coordinate reference point $P$ usually refers to the rotational center of the robot so that the coordinates remain constant when it rotates in place~\cite{multiwmr}.
In our case, it is the middle point between the centers of the two actuated wheels.

Note that in addition to the posture $q$, the complete configuration states of a mobile robot include the orientations of unfixed caster wheels and the rotation angles of all the wheels.
However, we consider only the states observable through internal and external sensors.
The remaining states are omitted in the formulation for simplicity.

\begin{figure*}[t]
\centering
\includegraphics[width=.99\textwidth]{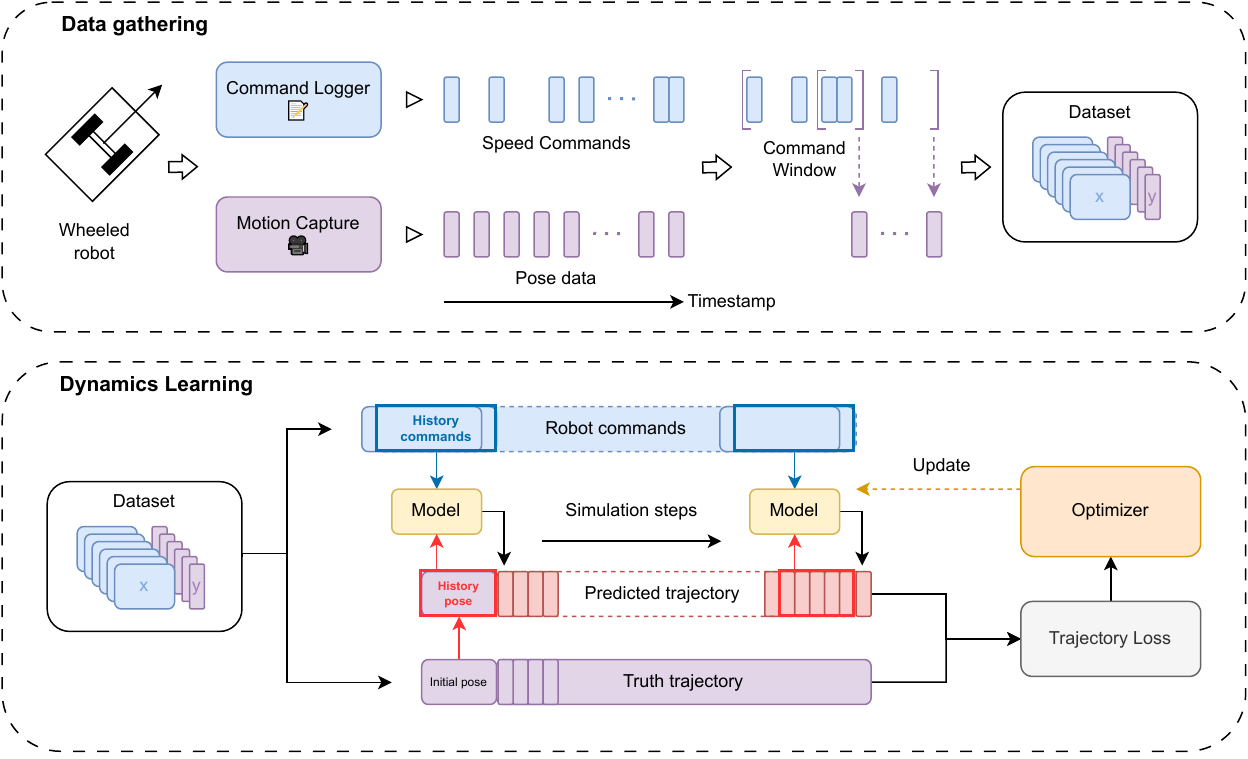}
\caption{
Overview of the data-driven simulation pipeline in D4W.
Two framework phases are shown: 1) data gathering: recording the robot poses with input command windows to form the dataset 2) dynamics learning: Evaluating and training the model with a sliding command and pose window along simulation steps.
}
\label{fig:_overview}
\end{figure*}

\paragraph{Robot control.}
We assume the robot uses conventional wheels, which cannot move sideways (nonslip condition).
The kinematics model of the unicycle-type WMR is therefore given as:
\begin{equation}
\begin{cases}
\dot{x} = s \cos{\theta}\\
\dot{y} = s \sin{\theta}\\
\dot{\theta} = \omega\\
\end{cases}
\end{equation}
Here, $\dot{x}, \dot{y}$, and $\dot{\theta}$ refer to the derivatives of the variables w.r.t time.
$s$ stands for the scalar velocity of the robot chassis in the longitudinal direction, and $\omega$ is its rotational speed around the coordinate reference point $P$.

Under the pure rolling condition, which assumes the contact point between each driving wheel and the ground has zero velocity, we can derive the following equations on the driving wheel velocity:
\begin{equation}
\begin{cases}
\displaystyle w_l = \frac{s - \omega R}{r}\\[1em]
\displaystyle w_r = \frac{s + \omega R}{r}\\
\end{cases}
\end{equation}
with $w_l$ and $w_r$ the angular velocities of the left and the right wheels, $R$ the distance from the middle point to each wheel, and $r$ the wheel radius.
In practice, however, the wheel contact points may slide in the longitudinal direction, making the actual wheel velocity $\hat{w_l}, \hat{w_r}$ independent from
the chassis velocity $s$.

To control the robot, commands are issued in the form of desired forward (backward) speed $s^c$ and angular velocity $\omega^c$ of the robot cart.
The internal speed controller and drive actuator convert the cart speed command to the rotation speed of each driving wheel and adjust the power output to match the result.

The dynamics model in a simulator estimates the robot's configuration states in each time step, given the input command and previously predicted ones.


\section{Methodology}
\subsection{Data-Driven Simulation Pipeline}

Analytical simulators achieve good explainability but lack alignment with reality\rej{if untuned}.
In D4W, we use data-driven approaches to compensate for real-world measurement data inaccuracies. Figure~\ref{fig:_overview} shows the overall workflow in D4W.

The first step is to gather the training data containing input commands and observed states in recorded robot trajectories.
The latter is used to infer the pose of the robot, creating the data set:
\begin{equation}
    \mathcal{D}=\{t^c_i, s^c_i, \omega^c_{i}\}_m \cup \{t_j, q_j\}_n~,
\end{equation}
where $t^{c}_{i}$ and $t_j$ indicate the timestamp of the command and pose data since they are gathered at different intervals in general cases.
We assume the timestamps are generated from reliable and synchronized time sources running at equal speeds.

The command sequence is then fed to the data-driven simulator to obtain the estimated trajectory.
The process starts by selecting an observed pose $q_{k}$ as the initial pose before the predictions.
\begin{equation}
\hat{q}_{0} = q_{k}~.
\end{equation}

Then, for each step $i$, the dynamics model $\mathcal{F}$ is queried with previous poses and input commands for the next output.
\begin{equation}
\label{eq:_sequential}
    \begin{aligned}
Q_i &= \{\hat{q}_j | \forall j, i - H < j \le i\}\\
C_i &= \{(t^c_j, s^c_j, \omega^c_j) | \forall j, t_{k+i} - T \le t^c_j \le t_{k+i}\}\\
\hat{q}_{i+1} &= \mathcal{F}(Q_i, C_i)~.\\
    \end{aligned}
\end{equation}
The pose set $Q_i$ keeps track of the robot poses estimated in previous $H$ simulation steps.
The command set $C_i$ is defined as the commands the robot received within a time window of length $T$ up to the current timestamp.
This ensures the model has constant input dimensions while retaining all the relevant information.
Finally, the model takes the command set $C_i$ and the history $Q_i$ as inputs and computes the next pose estimate $\hat{q}_{i+1}$.

Since the data is serial in the time dimension, a simulated trajectory $\hat{T}=\{\hat{q}\}$ can be computed for any contiguous subsequence $T=\{q_{k+i} | 0 < i \le s \}$ of the actual pose data.
Then, we can update the model parameters by optimizing the error between the simulation and the ground truth, for example, the MSE loss:
\begin{equation}
\label{eq:_mse}
\mathcal{L}=\frac{1}{n} \sum (q_{k+i} - \hat{q}_i)^2~.
\end{equation}
Depending on the model type, we can choose different optimization methods, e.g., gradient descent methods~\cite{shac, Zamora2021PODSPO} for differentiable models or Bayesian optimization~\cite{skopt} for non-differentiable ones.

\begin{figure}[t]
\centering
\includegraphics[width=.49\textwidth]{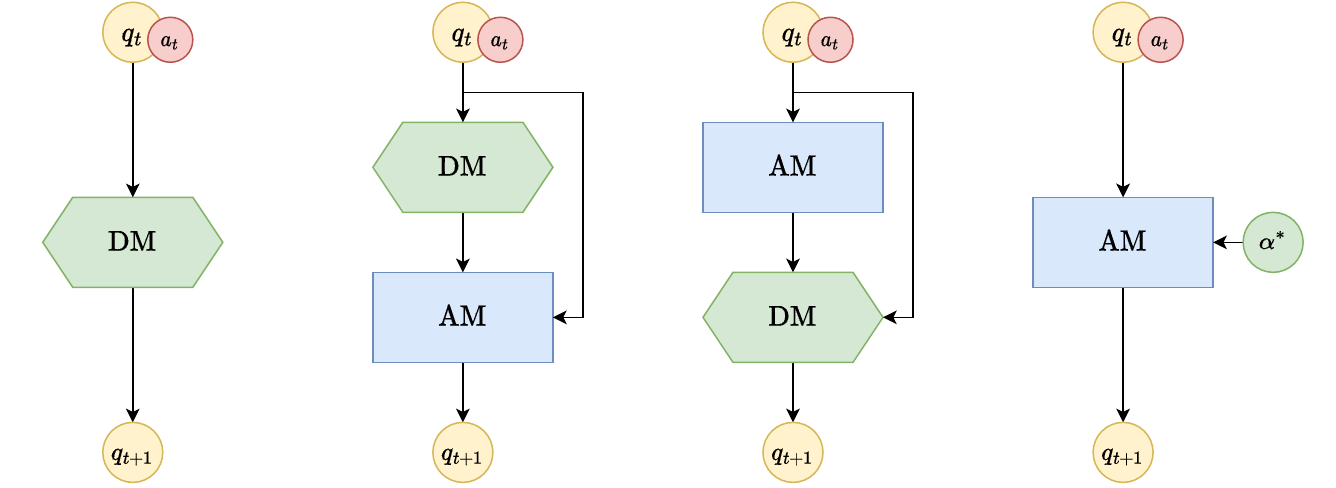}
\caption{
Possible architectures of the data-driven model.
From left to right: 1) pure data-driven model 2) dynamical hybrid model 3) kinematic hybrid model 4) analytical parameter model
}
\label{fig:_models}
\end{figure}

\subsection{Simulation Models}

We propose a set of architectural designs for the models inside the D4W simulator instead of the analytical ones.
In this section, $\operatorname{AM}$ refers to the analytical model, $\operatorname{DM}$ refers to the learnable data-driven model, $(*)$ includes all the inputs to the model, and $\alpha$ denotes model parameters.

\subsubsection{Pure Data-driven Models}

This method uses a neural network to directly learn the mapping from the robot inputs to its trajectory points~\cite{gnnphys, interphys, particledyn, contactmodel}.
\begin{equation}
\mathcal{F}(*)=\operatorname{DM}(*)
\end{equation}

Since the training data is sequential in time, recurrent networks~\cite{lstm, rnn} can be used to capture the temporal relations between robot poses.
The model can be optimized by computing the gradient of the loss and backward propagating it through time (BPTT)~\cite{brax}.

\subsubsection{Dynamical Hybrid Models}
\label{sss:_dyn_models}

Instead of end-to-end learned models, we can reuse and transform existing analytical models into data-driven ones by learning the corrected dynamical input in the form of additional forces and torques.
The dynamical correction represents physical interactions unanticipated in the original simulation, such as non-pure rolling friction.
The analytical model incorporates the learned forces into the computation process and derives the robot posture as the output.
\begin{equation}
\mathcal{F}(*)=\operatorname{AM}(\operatorname{DM}(*), *)
\end{equation}

The differentiability of the whole model depends on the implementation of the analytical simulator.
Unfortunately, most simulators in practical robot research and deployments do not support differentiable programming~\cite{gazebo, ode,ros}, making gradient-based optimization impossible for this model.

\subsubsection{Kinematic Hybrid Models}
\label{sss:_kine_models}

Another approach is to learn a correction on the analytical simulation results, namely the estimated position and orientation of the robot and their time derivatives.
\begin{equation}
\mathcal{F}(*)=\operatorname{DM}(\operatorname{AM}(*), *)
\end{equation}

The analytical part of the model does not block gradient flow in this case.
Thus, the learnable components are still differentiable.
Depending on the design, the two components can be parallel (residual) or sequential.

\subsubsection{Analytical Parameter Model}

This method differs from the above types in that it learns the initial configuration parameters of the analytical model while leaving the simulation process completely unmodified.
\begin{equation}
\begin{gathered}
\alpha^*_{AM} = \argmin_{\alpha} \sum_{i} \mathcal{L}(\operatorname{AM}(*, \alpha), q_i)\\
\mathcal{F}(*) = \operatorname{AM}(*, \alpha^*_{AM})\\
\end{gathered}
\end{equation}

The hyperparameter tuning~\cite{modnas, skopt} approaches suit this scenario if gradient descent is unavailable.


\begin{figure*}[htb]
\centering
\begin{subfigure}[b]{0.325\textwidth}
    \centering
    \includegraphics[width=\textwidth]{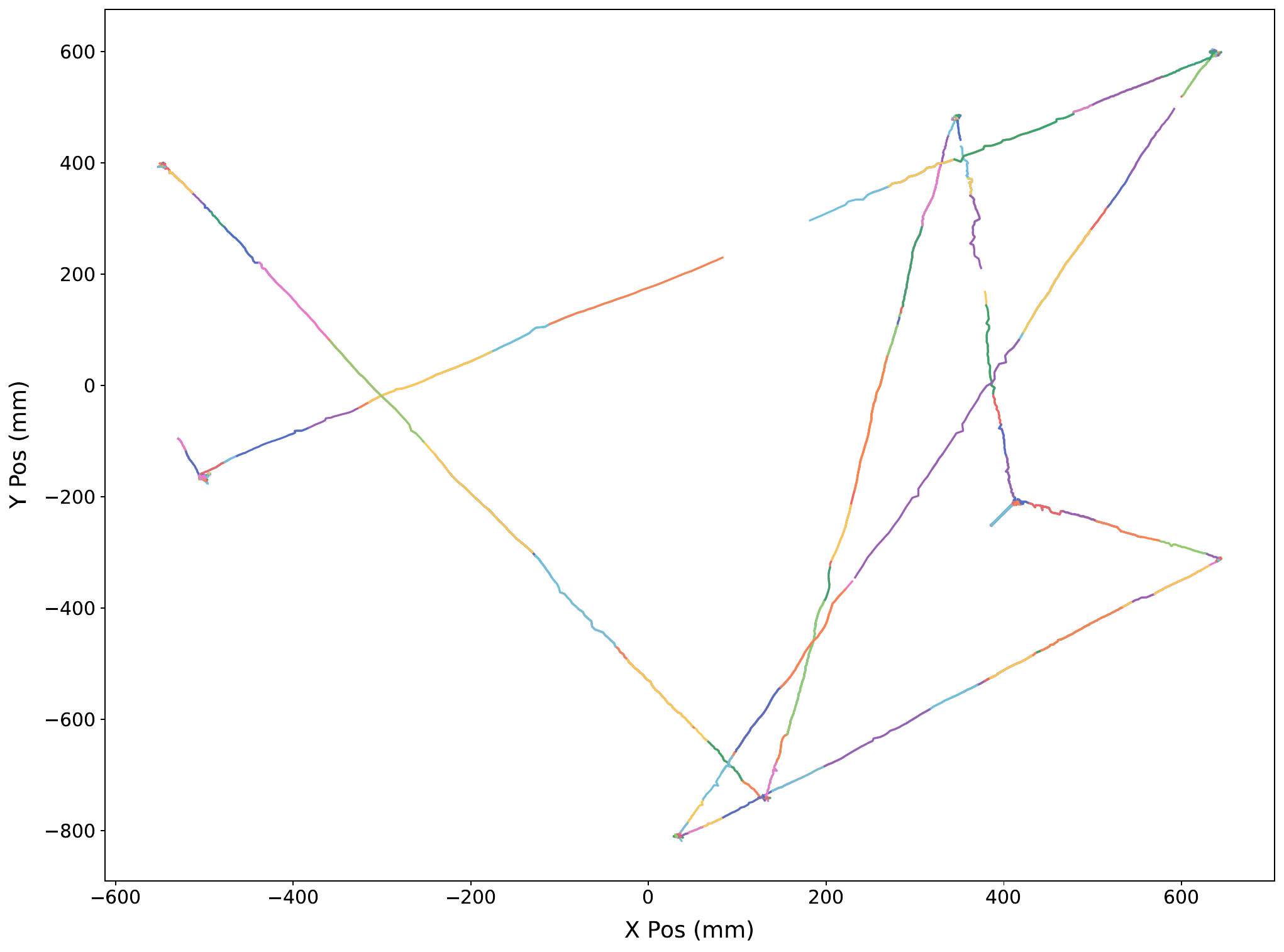}
    \caption{No transformations}
\end{subfigure}
\begin{subfigure}[b]{0.33\textwidth}
    \centering
    \includegraphics[width=\textwidth]{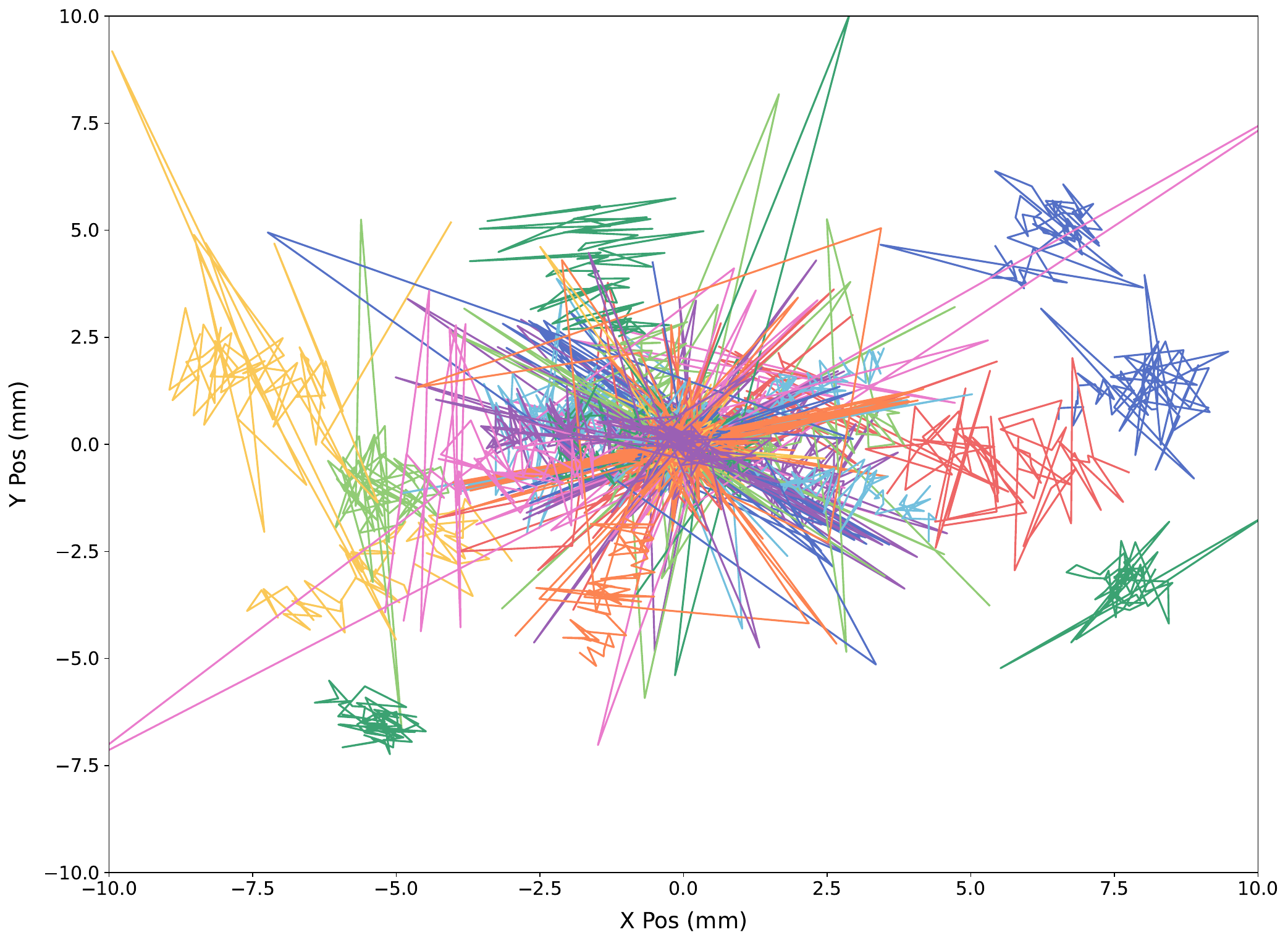}
    \caption{Translational only}
\end{subfigure}
\begin{subfigure}[b]{0.325\textwidth}
    \centering
    \includegraphics[width=\textwidth]{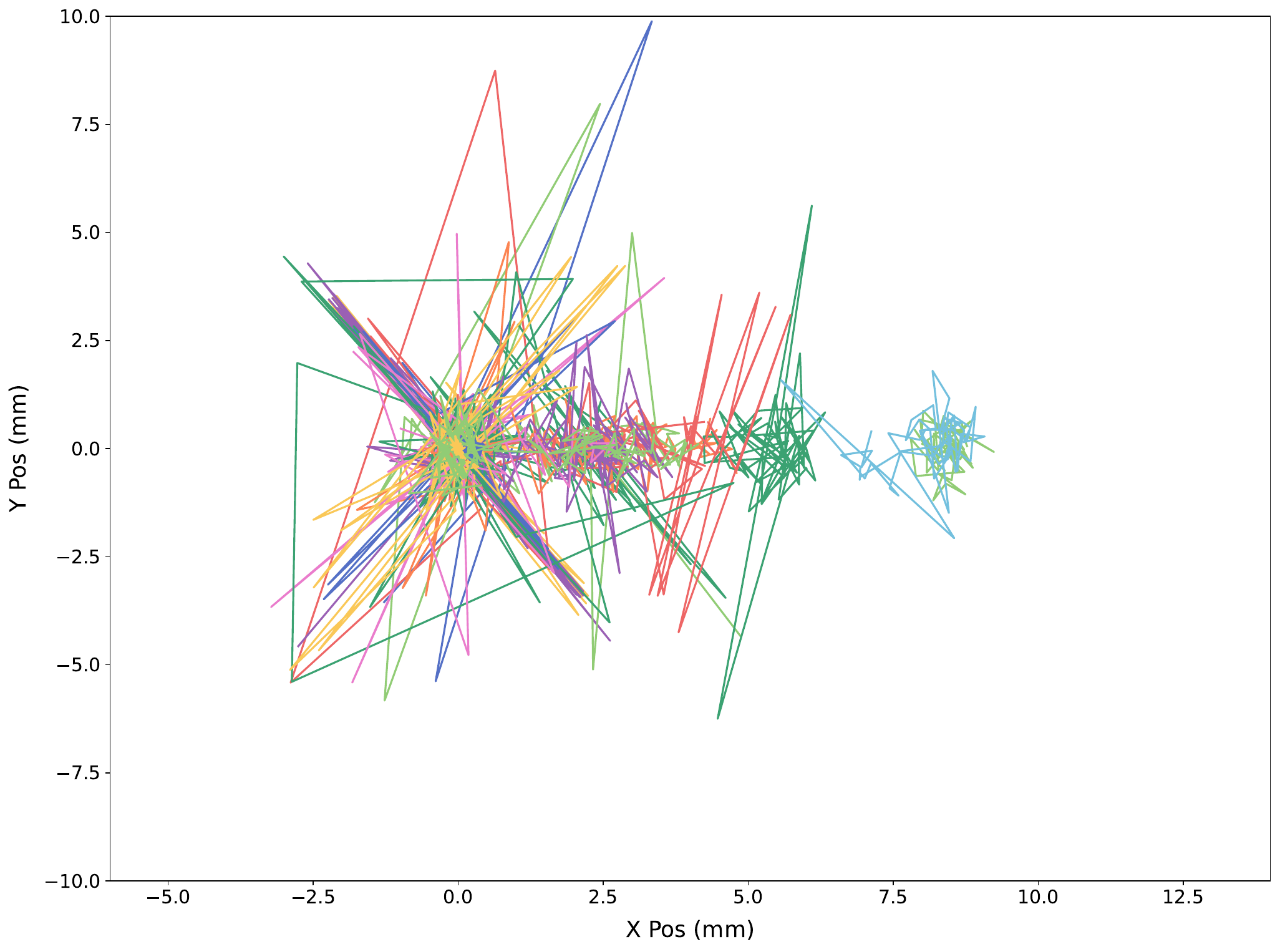}
    \caption{Translational \& rotational (Egocentric)}
\end{subfigure}
    \caption{
Training trajectories under various transformations.
Egocentric transformation yields an easier training target.
}
    \Description{.}
    \label{fig:_ego_transforms}
\end{figure*}

\subsection{Egocentric Simulation}
\label{ss:_ego_sim}

As stated in the above formulations, the data-driven model takes previously estimated poses as inputs for the next prediction due to the pose data's temporal dependency and sequential nature.
However, the poses recorded in the data set are based on global coordinates, which results in learned dynamics that do not respect translational and rotational symmetry if trained on such data.

A dynamical system satisfies translational symmetry if its behavior is invariant under any translation (shift).
A similar definition exists for rotational symmetry and time translational symmetry.
These symmetries hold for most physical systems~\cite{noether}, including those under wheeled robot settings.

To ensure the learned dynamics conform to the symmetries, we perform several transformations on both the model's input and output poses during each prediction so that the model is only aware of the localized motions within the history window.

In detail, a global offset vector keeps track of the transformation from the coordinates in the egocentric frame to the original data frame.
The offset vector is set as the initial pose at the beginning of a simulation trajectory in Eq.~\ref{eq:_sequential}.
\begin{equation}
\begin{gathered}
\Delta q = (\Delta x, \Delta y,  \Delta \theta)^T\\
\Delta q_0 = q_{k}
\end{gathered}
\end{equation}

After each simulation step, the model output is transformed back to the global frame, and the offset is updated to keep the historical poses localized.
\begin{equation}
\begin{aligned}
\hat{r}_{t+1} &= \mathcal{F}(*)\\
\hat{q}_{t+1} &= \Delta q_t + R\cdot\hat{r}_{t+1} \\
\Delta q_{t+1} &= \Delta q_t + R\cdot\hat{r}_{t-H}\\
\end{aligned}
\end{equation}

Here, $\hat{r}$ is the predicted pose in the localized frame. $R$ represents the transformation matrix applied on $\hat{r}$, which rotates the pose according to $\Delta \theta$ and accumulates the orientation change.
\begin{equation}
R=\begin{bmatrix}  
    \cos{\Delta \theta} & -\sin{\Delta \theta} & 0 \\  
    \sin{\Delta \theta} & \cos{\Delta \theta} & 0 \\  
    0 & 0 & 1 \\  
\end{bmatrix}
\end{equation}

When training the dynamics model, the loss function compares the global pose of each step $\hat{q}_t$ against the observed global pose $q_t$.
\begin{equation}
\begin{aligned}
\operatorname{Loss}_{t+1}&=\mathcal{L}(\hat{q}_{t+1}, q_{t+1})\\
&=\mathcal{L}(\Delta q_{t} + R\cdot\hat{r}_{t+1}, q_{t+1})\\
\end{aligned}
\end{equation}

According to the chain rule, the gradient of the model parameters $\alpha$ is
\begin{equation}
\begin{aligned}
\parfrac{\mathcal{L}}{\alpha} &= \chainrule{\mathcal{L}}{\hat{q}}{\alpha}\\
&=\parfrac{\mathcal{L}}{\hat{q}} \cdot \Big(\chainrule{\hat{q}}{\hat{r}}{\alpha} + \chainrule{\hat{q}}{\Delta q}{\alpha}\Big)\\
&=\parfrac{\mathcal{L}}{\hat{q}} \cdot \Big(R\parfrac{\hat{r}}{\alpha} + \chainrule{\hat{q}}{\Delta q}{\alpha}\Big)\\
\end{aligned}
\end{equation}

The derivation shows that the transformation matrix $R$ appears in the gradient computation, complicating the outcome and negatively affecting convergence.
Therefore, we propose to compute the loss function in the egocentric frame by converting the ground truth pose instead:
\begin{equation}
\begin{aligned}
r_{t+1} = R^{-1}(q_{t+1} - \Delta q_{t}),
\end{aligned}
\end{equation}
where $R^{-1}$ is the inverse of the transformation matrix equivalent to rotating the pose in the opposite direction.
Then we have
\begin{equation}
\begin{aligned}
\operatorname{EgoLoss}_{t+1}&=\mathcal{L}(\hat{r}_{t+1}, r_{t+1})\\
&=\mathcal{L}(\hat{r}_{t+1}, R^{-1}(q_{t+1} - \Delta q_{t}))\\
\end{aligned}
\end{equation}
and
the gradient of $\alpha$ now becomes
\begin{equation}
\begin{aligned}
\parfrac{\mathcal{L}}{\alpha} &= \chainrule{\mathcal{L}}{\hat{r}}{\alpha}\\
&=\parfrac{\mathcal{L}}{\hat{r}} \cdot (\parfrac{\hat{r}}{\alpha} + \chainrule{\hat{r}}{\Delta q}{\alpha}),
\end{aligned}
\end{equation}
which does not contain $R$.
Note that $R$ still exists implicitly in the derivatives of $\Delta q$, which are the BPTT terms and can be modulated by truncating the gradient flow.

Alternatively, if only translational symmetry is enforced, we have $\Delta \theta=0$ throughout the trajectory, and $R$ is the identity matrix.
This further reduces the number of non-linear operators in the computational graph, which is expected to help alleviate the problem of complex loss landscape caused by the long BP chain through time.

As for the time translation symmetry, we recalculate the timestamps of input commands in each step such that the current simulation time is always at the origin, removing the singular absolute epoch in the raw data.
The timestamps of the pose data are not modified since they are used as time origins and do not appear in the simulation process.

We visualize the transformations employed in~Figure~\ref{fig:_ego_transforms}.

\subsection{Implementation Choices}

\subsubsection{Loss Functions}

We consider alternative loss terms in place of the standard step-wise MSE loss in Eq.~\ref{eq:_mse}.
Inspired by the computer vision and imitation learning literature~\cite{chen2022imitation}, we introduce the $\text{Chamfer-}\alpha$ loss that compares the simulation outcomes trajectory-wise. The loss function is originally defined as follows:
\begin{equation}
\text{Chamfer-}\alpha(A, B) = \frac{1}{|A|} \sum_{a \in A} \min_{b \in B} \|a - b\|_2^\alpha + \frac{1}{|B|} \sum_{b \in B} \min_{a \in A} \|a - b\|_2^\alpha  
\end{equation}

It measures the dissimilarity between two sets of points $A$ and $B$ by computing the average of the Euclidean distances between each point in one set and its nearest neighbor in the other set, and vice versa.
The parameter $\alpha$ controls the weight or sensitivity of the loss function.

Under our settings, the loss function can be transformed as follows:
\begin{equation}
\mathcal{L}= (1-\alpha) \frac{1}{|T_k|} \sum_{q \in T_k} \min_{\hat{q} \in \hat{T}_k} (q - \hat{q})^2 + \alpha \frac{1}{|\hat{T}_k|} \sum_{\hat{q} \in \hat{T}_k} \min_{q \in T_k} (\hat{q} - q)^2
\end{equation}

Minimizing the first part of the loss allows the learned trajectory to match the reference one on a global scale because it covers every ground truth point.
Conversely, the second loss term measures the local deviation from the actual data in each predicted point.
$\alpha$ is the weighting factor between the global and local loss.
This helps faster convergence by relaxing the step-by-step MSE loss at the expense of potential performance drop.
As another option, points can be skipped when calculating the loss, resulting in a gapped loss function.

We also employ L2 regularization for gradient-based methods to sparsify the model parameters.
Gradients are optionally normalized or wholly cut off at intervals for better stability and faster convergence.

\subsubsection{Progressive Training}

Intuitively, the difficulty of accurately simulating a trajectory drastically increases with the trajectory length due to compounding errors.
On the other hand, the learned dynamics can be considered practical and dependable only if it performs well for a sufficiently long period without deviating too far from the truth.

To solve this problem, we propose to train the model progressively on longer episodes.
To reduce the total training time, the sequence length is increased exponentially rather than linearly.

\subsubsection{Unattended Data Collection}
\label{sss:_data_collect}

To obtain the training data efficiently and improve the usability of the D4W framework, we automate the process of gathering observations in physical experiments by giving out a sequence of randomly sampled commands to the robot and monitoring the execution status.

\begin{figure}[htb]
\centering
\includegraphics[width=.45\textwidth]{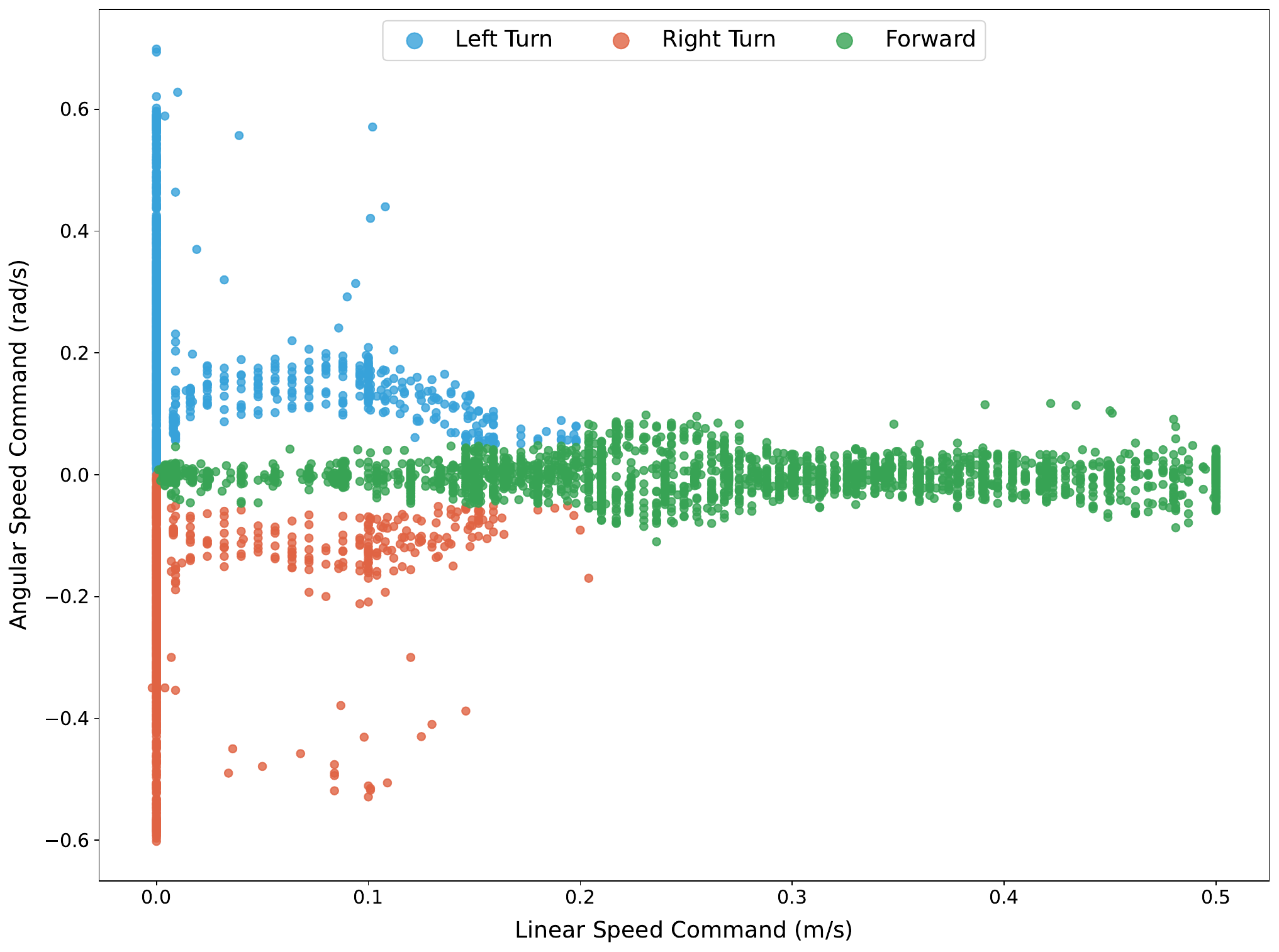}
\caption{Robot commands gathered in data collection.
The X and Y axes show the longitudinal and the angular speed commands.
Commands are clustered into three types: forward, left turn, and right turn.
}
\label{fig:_cmd_dist}
\end{figure}

Specifically, the robot repeatedly performs the following steps: 1) Uniformly sample a valid target position and orientation in a predefined safe area. 2) The navigation module issues speed commands to reach that position. 3) The robot pose is monitored and will interrupt the command if it reaches the target or is stuck for an extended period. 4) The target and the speed command are reset.
We visualize the distribution of collected speed commands in Figure~\ref{fig:_cmd_dist}.

\section{Evaluations}

This section presents the evaluations conducted to assess the performance of the analytical baselines, pure data-driven, and hybrid models.
Additionally, we perform an ablation study to analyze the impact of specific components and techniques used in this work.

\subsection{Dataset Building}

\begin{figure*}[htb]
\centering
\begin{subfigure}[b]{0.32\textwidth}
    \centering
    \includegraphics[width=\textwidth]{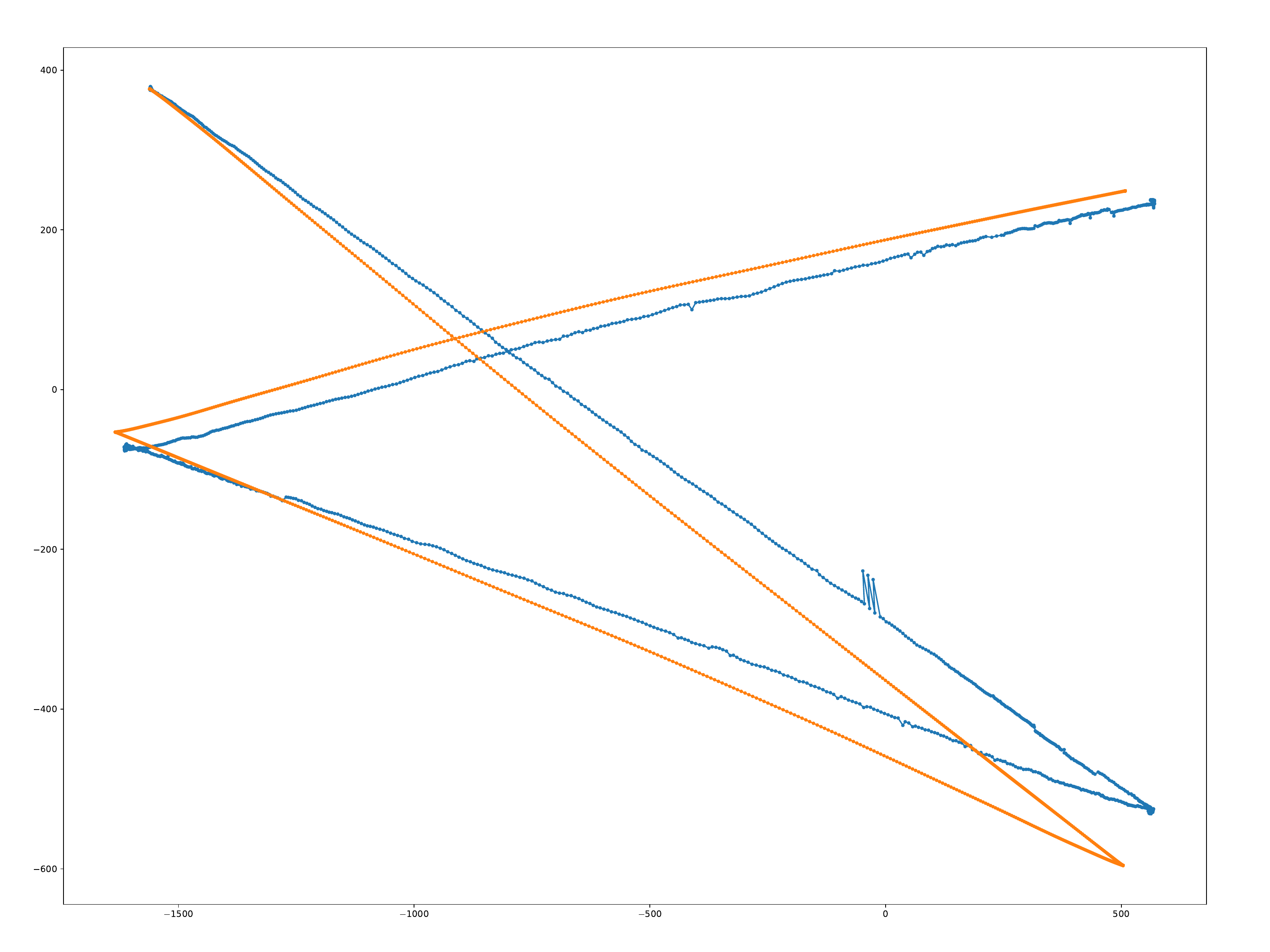}
    \caption{X-Y Trajectory}
\end{subfigure}
\begin{subfigure}[b]{0.32\textwidth}
    \centering
    \includegraphics[width=\textwidth]{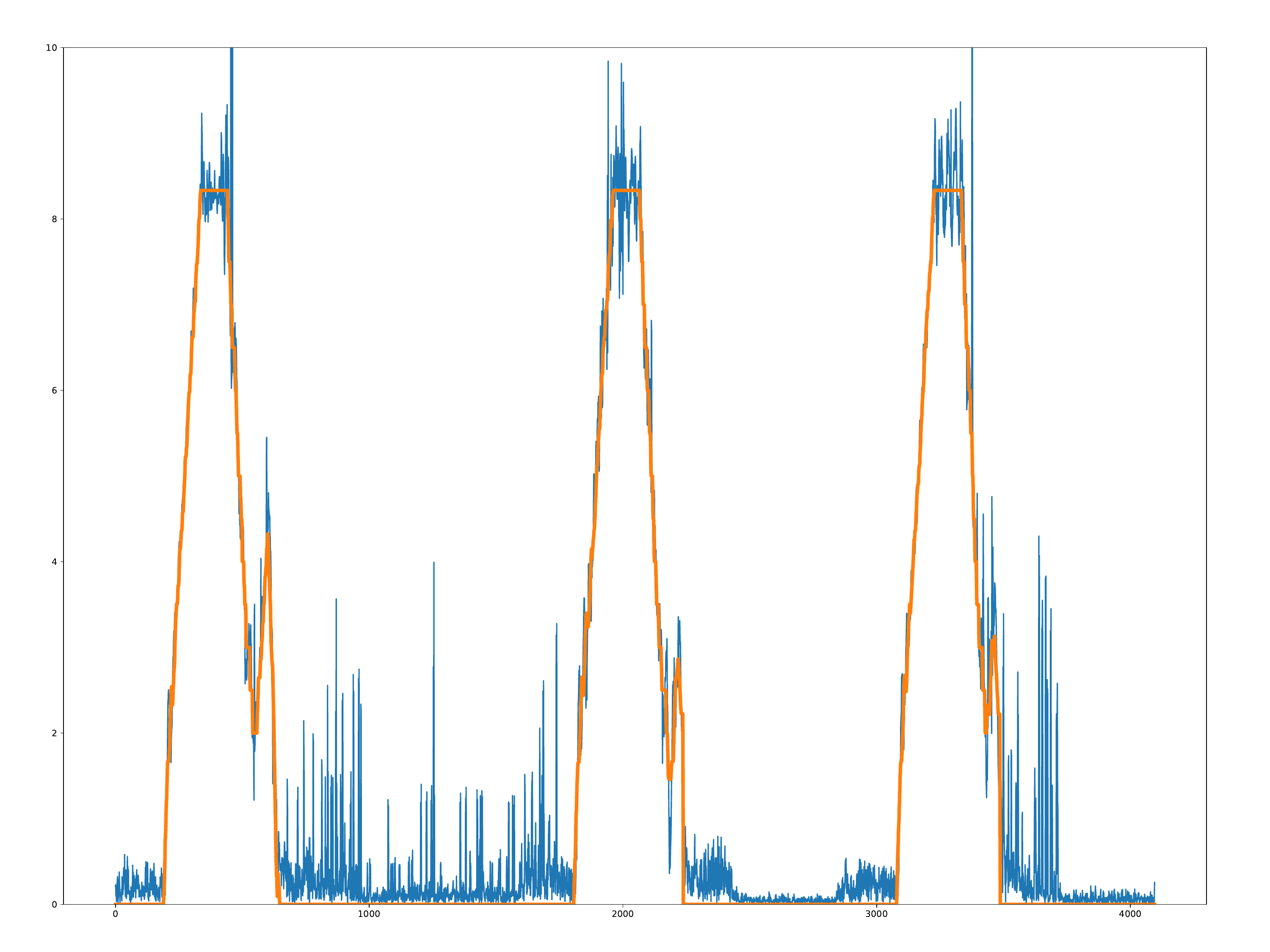}
    \caption{Position delta}
\end{subfigure}
\begin{subfigure}[b]{0.32\textwidth}
    \centering
    \includegraphics[width=\textwidth]{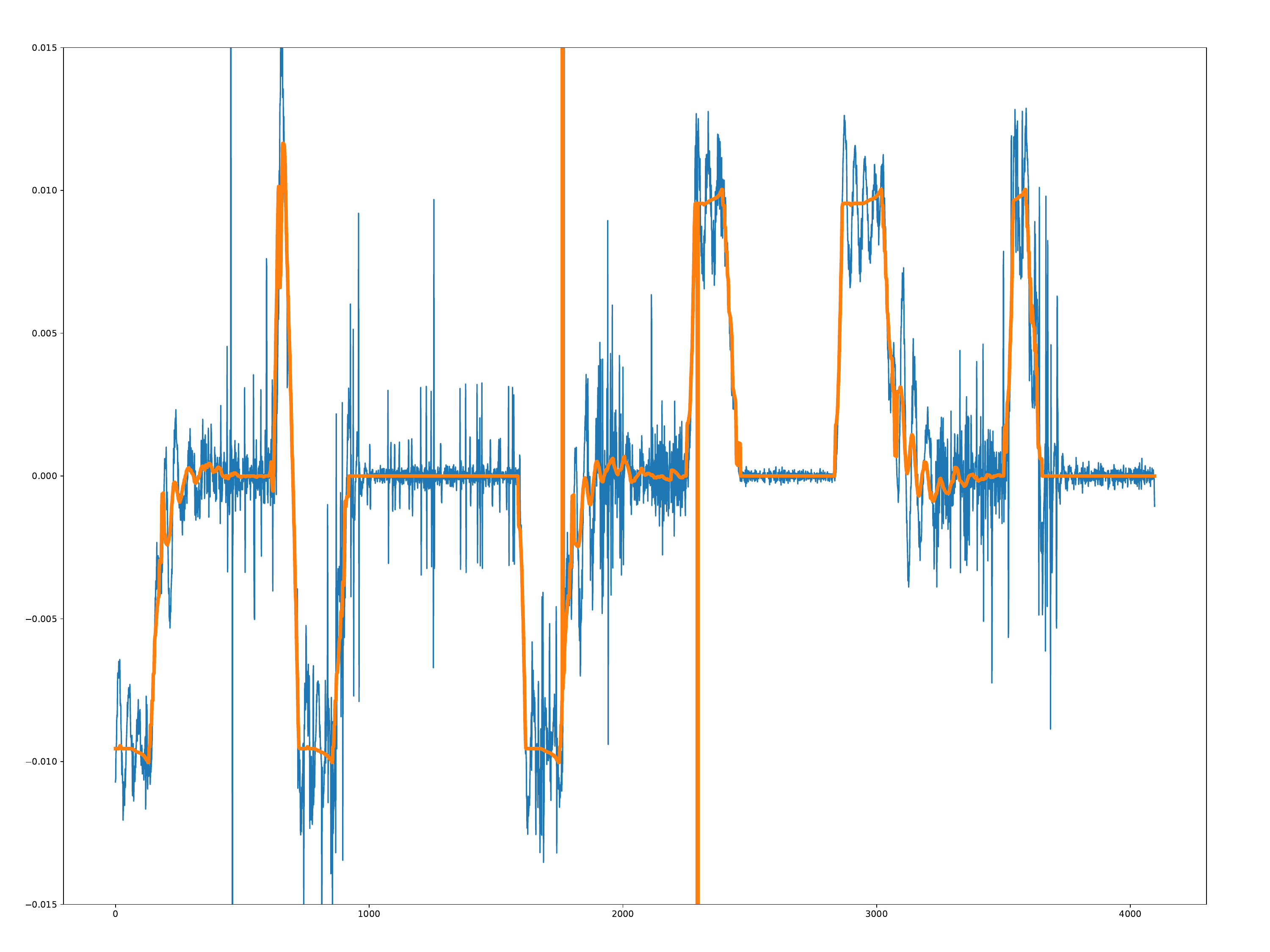}
    \caption{Orientation delta}
\end{subfigure}
    \caption{
Comparison between a ground truth trajectory and one predicted by the handmade formula (Blue: ground truth; Orange: predicted).
}
    \Description{.}
    \label{fig:_eval_traj_pred}
\end{figure*}

We choose an experimental service robot model used in warehouse logistic applications for evaluations.
A motion capture system records the training pose data while the robot follows the automated data collection procedure in~subsubsection~\ref{sss:_data_collect}.
The capture system reports the position and rotation of the robot chassis at 60 fps with a claimed millimeter-level error.
However, there is still a significant amount of noise in the data, as shown in~Figure~\ref{fig:_eval_traj_pred}.

We log the pose and command data simultaneously for 2000 seconds, acquiring 120k and 50k valid frames of each type of data.
Overlapping between training trajectories is allowed to augment the training dataset. 

The gathered poses are partially visualized in Figure~\ref{fig:_eval_dset}.
Note that the coordinates in the data belong to a different point in the robot chassis other than the reference point $P$, and corrections are done according to the known form factors.

For the following evaluations, 30 \% of the data is held out as the test set, and the rest is used for training each model.
The pose history and input window sizes are set to 1 and 200 ms, respectively.

\subsection{Analytical Baselines}

We start by evaluating analytical baselines, which provide a reference for comparing the performance of data-driven models.

\subsubsection{Experiment Details}

We use Isaac Sim to build the physical simulation environment.
The robot model is imported from its original design schematics, containing precise physical and geometrical information on the product.
Simplified shapes like cylinders and spheres are used as colliders in place of the visual mesh to improve simulation stability.
A built-in differential controller actuates the robot's drive wheels.
To compare the results with the ground truth, we transform the simulator's internal reference frame to the dataset frame similar to~subsection~\ref{ss:_ego_sim}.
The simulator runs 256 robot environments in parallel, which are reset at the beginning of each trajectory batch.

To enable gradient-based optimizations on hybrid models in subsequent experiments, we also provide a handmade formula written in PyTorch~\cite{pytorch} as a differentiable analytical model.

\subsubsection{Results and Discussions}

We present the results on the analytical baselines in Table~\ref{tab:_eval_models}.
We report performance metrics as step average RMSE in robot positions evaluated under different trajectory lengths.
The orientation error is excluded because it is often negligible compared to positional errors.
The resulting metric is also the average distance deviated in each frame.

\begin{table*}[ht]
\caption{Results on analytical and data-driven models (RMSE error)}
\label{tab:_eval_models}
\begin{tabular}{cccccccc}
  \toprule
  \multirow{2}*{Dynamical Models} & Trajectory lengths & \multicolumn{6}{c}{ Trajectory lengths (Evaluation)}\\
  \cmidrule{3-8}
  & (Training) & 1 & 8 & 64 & 512 & 4096 & 32768 \\
  \midrule
  Isaac Sim & 0 &  2.9479 & 4.5208  & 7.8473 & 66.893 & 424.33 & 1207.9 \\
  Hand-Formulated & 0 & 2.0533 & 2.4800 & 3.6603 & 12.201 & 49.947 & 133.37 \\
  \midrule
  LR & 1 & 1.9609 & 9.8257 & 74.006 & 426.96 & 1012.89 & 14503 \\
  LR & 64 & 1.8531 & 2.4364 & 3.7029 & 63.136 & 1192.4 & 3403.6\\
  LR & 4096 & 1.8532 & 2.4363 & \textbf{3.6235} & 112.62 & 1385.9 & 2315.5 \\
  MLP & 1 & 1.8525 & 2.4270 & 4.1211 & 374.22 & 2593.0 & 6026.3\\
  MLP & 64 & 1.8527 & 2.4283 & 3.6551 & 283.90 & 2576.5 & 9183.7\\
  MLP & 4096 & \textbf{1.8523} & 2.4224 & 3.8515 & 38.461 & 224.89 & 1202.6 \\
  \midrule
  Formulated + MLP & 1 & 1.8541 & 2.4444 & 3.9437 & 320.82 &  2662.8 & 19210 \\
  Formulated + MLP & 64 & 1.8541 & 2.4445 & 3.7923 & 34.024 & 327.64 & 1104.4 \\
  \textbf{Formulated + MLP} & 4096 & 1.9121 & 2.2210 & 3.8150 & \textbf{10.616} & \textbf{33.025} & \textbf{72.102} \\
  MLP + Formulated & 64 & 30.055 & 129.8 & 212.08 & 492.56 & 1196.0 & 1476.7 \\
  MLP + Formulated & 4096 & 11.421 & 41.617 & 69.72 & 488.47 & 1159.4 & 1392.3 \\
  \bottomrule
\end{tabular}
\end{table*}

\subsection{Data-driven Models}

In this section, we evaluate data-driven models of various architectures and compare them with the baselines.

\subsubsection{Experiment Details}

For pure data-driven models, we choose Linear Regression (LR) and MLP to represent feed-forward neural networks.
Recurrent networks have been reported to have poor performance~\cite{walkinmin} and are omitted in the evaluation.
To ensure differentiability, we combine the learnable components with the formulated dynamics for hybrid models.
The training is done progressively with increasing trajectory lengths until the maximum is reached.


The code is implemented with Pytorch~\cite{pytorch} v1.9.0.
Each model is trained using a single NVIDIA Tesla V100-SXM2-32GB GPU.

\subsubsection{Results and Discussions}

For better comparison, we merge the data-driven model results into Table~\ref{tab:_eval_models}.
Specifically,  Formulated + MLP refers to the dynamical hybrid model in~subsubsection~\ref{sss:_dyn_models}, and MLP + Formulated refers to the kinematic hybrid model in~subsubsection~\ref{sss:_kine_models}.

Several insights are:

\begin{itemize}[leftmargin=15pt]
    \item Data-driven models generalize better than analytical ones on long trajectories.
    \item The differentiable hybrid model has the best prediction accuracy in a relatively short training time.
    \item Models trained on short trajectories tend to fail on longer ones, while training on longer ones generalizes better on all lengths.
\end{itemize}

\subsection{Ablation Study}

In this section, we conduct an ablation study to analyze the individual contributions of specific components and techniques used in the D4W framework.

\subsubsection{Egocentric Transformation}

\begin{figure}[htb]
  \centering
      \includegraphics[width=.45\textwidth]{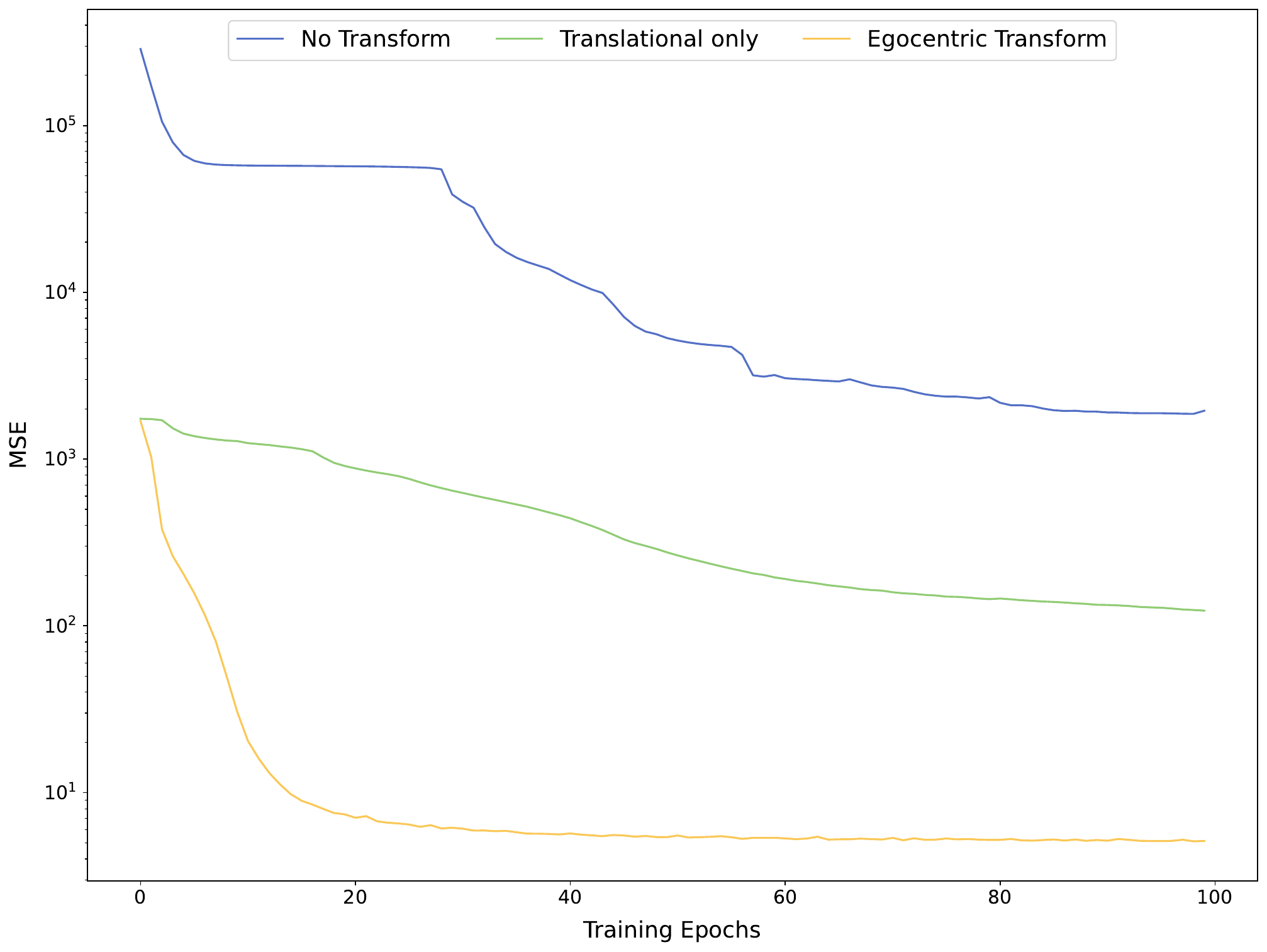}
      \caption{
Training progress under various transforms.
The model trained on egocentric-transformed data converges faster than those trained with other transforms.
}
      \Description{.}
      \label{fig:_eval_ego}
\end{figure}

In this subsection, we evaluate the impact of the egocentric transformation component.
We compare the training progress of a single linear layer (LR) obtained with and without the egocentric transformation.
Figure~\ref{fig:_eval_ego} shows the RMSE loss with training steps and Table~\ref{tab:_eval_ego} shows the final performance of models with different transforms.

The results show that egocentric transformation significantly improves generalization ability by observing fundamental symmetries in the learned dynamics.

\subsubsection{Loss function}

\begin{figure}[htb]
  \centering
      \includegraphics[width=.45\textwidth]{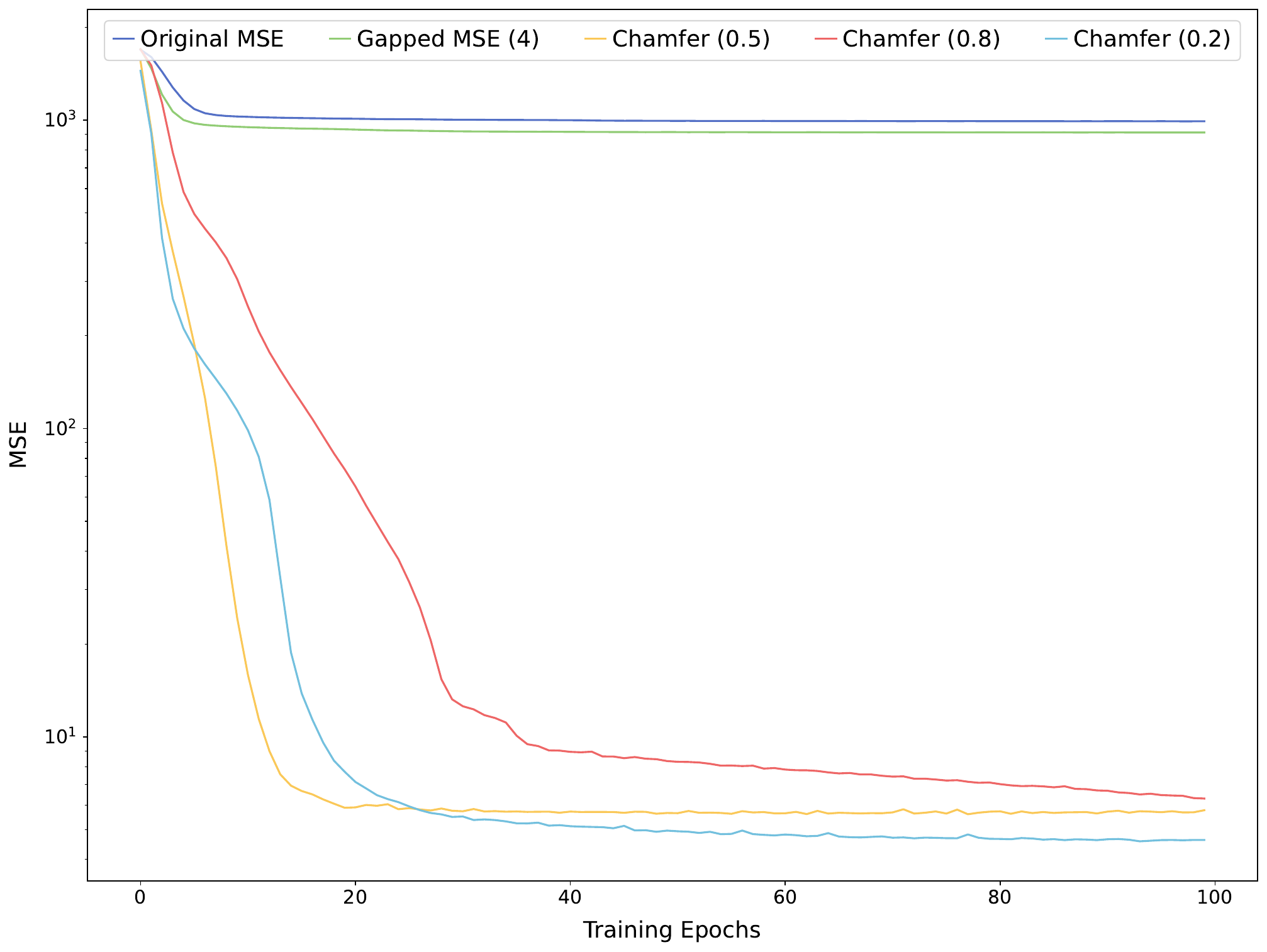}
      \caption{
Training progress (MSE error) with various loss functions.
The Chamfer loss contributes to faster convergence and improved final performance.
}
      \Description{.}
      \label{fig:_eval_loss}
\end{figure}

Here, we investigate the effect of different loss functions on the performance.
Using an MLP model as the optimization target, we evaluate multiple loss functions and compare their results in~Figure~\ref{fig:_eval_loss}.
We evaluate the full RMSE error on the training trajectories of length 64 while using different loss functions as objectives.

Due to its relaxed dissimilarity measure, the Chamfer loss contributes to faster convergence than MSE variants.
However, the computation complexity is quadratic with the sequence length, leading to noticeably reduced training speed on larger data sets.
A potential enhancement exploiting the sequentially of the data is to limit the nearest neighbor comparison to K adjacent elements in the trajectory.

\begin{table}[ht]
\caption{Results on data transformations}
\label{tab:_eval_ego}
\begin{tabular}{ccccccc}
  \toprule
  \multirow{2}*{Transforms} & \multicolumn{6}{c}{Trajectory lengths} \\
  \cmidrule{2-7}
  & 1 & 8 & 64 & 512 & 4096 & 32768 \\
  \midrule
  None & 7.0106 & 17.262 & 87.658 & 473.31 & 2073.8 & 8379.1 \\
  Translational & 7.0987 & 9.2082 & 18.827 & 535.97 & 1850.0 & 3770.9 \\
  \textbf{Egocentric} & 1.8533 & 2.4355 & 3.6303 & 427.81 & 1125.5 & 1061.6 \\
  \bottomrule
\end{tabular}
\end{table}




\section{Related Work}

\subsection{Analytical Simulators}

Analytical simulators have been widely used in robot research and development.
These simulators rely on mathematical models and equations to simulate the robot dynamics.


Isaac Sim~\cite{orbit} is a GPU-accelerated simulation platform built on top of physX~\cite{physx} that offers a rich set of pre-built robot models, environments, and sensors.
It provides a realistic and customizable simulation environment for testing and validating various robotic applications, including perception, planning, and control.

Analytical models often assume ideal conditions and simplified dynamics~\cite{comparesim}, which may not accurately capture the complexities of real-world systems.

\subsection{Differentiable and Learnable Simulators}

Differentiable simulators have gained wider attention recently due to their ability to directly leverage data-driven techniques and optimize control algorithms~\cite{e2ediffctrl,chen2022imitation}.
Gradsim~\cite{gradsim} introduces end-to-end differentiability in simulation and rendering, allowing for learning dynamics from real-world interaction videos.
Brax~\cite{brax} proposes a highly scalable multi-body simulator in JAX~\cite{jax} to accelerate RL research.
Other works focus on differentiable dynamics of deformable objects~\cite{chen2023daxbench, disect, chainqueen, diffcloth}.

Some methods combine physics-based simulation with machine-learning techniques to learn accurate models of robot dynamics~\cite{flexphys, compdyns}.
DensePhysNet learns representations of physical objects for robot manipulation tasks.
NeuralSim implements a differentiable simulator for articulated rigid-body dynamics that introduces neural networks in the computation graph and evaluates on MuJoco.
On the other hand, intuitive physics models use graph neural networks (GNN) to model interactions between physical entities.

Learning environment dynamics is also investigated in reinforcement learning~\cite{Lai2020BidirectionalMP} and imitation learning~\cite{Wu2019ModelIF} literature as a sequential decision-making problem~\cite{Wen2023LargeSM}.

There exists research on data-driven methods specialized for wheeled robots.
\cite{Isenberg2023SystemIO} uses a motion capture system as observations to identify parameters in pure analytical models.
\cite{khan2022dynamic} and \cite{Uddin2020SystemIO} use neural networks to predict wheeled robot movement in simulated contexts without evaluations on actual robots.

\section{Conclusion}

In this paper, we introduce the D4W framework for dependable data-driven dynamics modeling of wheeled robots.
The framework combines physics-based simulation with data collection and machine learning techniques to learn accurate dynamics models while focusing on efficiency and generalizability by enforcing physical invariances on the learned dynamics.
It demonstrates superior performance through evaluation and comparative analysis compared to traditional analytical models.
Real-world validation experiments confirm the applicability and robustness of the learned dynamics models from D4W.

For the future work of D4W, we plan to provide interoperability for more simulation frameworks and extend the learning process to a broader range of robots.
As current experiments rely on an external motion capture system, another direction would be switching to onboard sources for robot pose data, for example, SLAM modules~\cite{Sturm2012ABF}.

\begin{acks}
This work is partially supported by National Key R\&D Program of China (2022ZD0114804) and National Natural Science Foundation of China (62076161). 
\end{acks}

\bibliographystyle{ACM-Reference-Format}
\bibliography{bib/bib3}


\begin{thebibliography}{44}


\ifx \showCODEN    \undefined \def \showCODEN     #1{\unskip}     \fi
\ifx \showDOI      \undefined \def \showDOI       #1{#1}\fi
\ifx \showISBNx    \undefined \def \showISBNx     #1{\unskip}     \fi
\ifx \showISBNxiii \undefined \def \showISBNxiii  #1{\unskip}     \fi
\ifx \showISSN     \undefined \def \showISSN      #1{\unskip}     \fi
\ifx \showLCCN     \undefined \def \showLCCN      #1{\unskip}     \fi
\ifx \shownote     \undefined \def \shownote      #1{#1}          \fi
\ifx \showarticletitle \undefined \def \showarticletitle #1{#1}   \fi
\ifx \showURL      \undefined \def \showURL       {\relax}        \fi
\providecommand\bibfield[2]{#2}
\providecommand\bibinfo[2]{#2}
\providecommand\natexlab[1]{#1}
\providecommand\showeprint[2][]{arXiv:#2}

\bibitem[Battaglia et~al\mbox{.}(2016)]%
        {interphys}
\bibfield{author}{\bibinfo{person}{Peter~W. Battaglia}, \bibinfo{person}{Razvan Pascanu}, \bibinfo{person}{Matthew Lai}, \bibinfo{person}{Danilo~Jimenez Rezende}, {and} \bibinfo{person}{Koray Kavukcuoglu}.} \bibinfo{year}{2016}\natexlab{}.
\newblock \showarticletitle{Interaction Networks for Learning about Objects, Relations and Physics}. In \bibinfo{booktitle}{\emph{NIPS}}.
\newblock
\urldef\tempurl%
\url{https://api.semanticscholar.org/CorpusID:2200675}
\showURL{%
\tempurl}


\bibitem[Bradbury et~al\mbox{.}(2018)]%
        {jax}
\bibfield{author}{\bibinfo{person}{James Bradbury}, \bibinfo{person}{Roy Frostig}, \bibinfo{person}{Peter Hawkins}, \bibinfo{person}{Matthew~James Johnson}, \bibinfo{person}{Chris Leary}, \bibinfo{person}{Dougal Maclaurin}, \bibinfo{person}{George Necula}, \bibinfo{person}{Adam Paszke}, \bibinfo{person}{Jake Vander{P}las}, \bibinfo{person}{Skye Wanderman-{M}ilne}, {and} \bibinfo{person}{Qiao Zhang}.} \bibinfo{year}{2018}\natexlab{}.
\newblock \bibinfo{booktitle}{\emph{{JAX}: composable transformations of {P}ython+{N}um{P}y programs}}.
\newblock
\urldef\tempurl%
\url{http://github.com/google/jax}
\showURL{%
\tempurl}


\bibitem[Chang et~al\mbox{.}(2016)]%
        {compdyns}
\bibfield{author}{\bibinfo{person}{Michael~B Chang}, \bibinfo{person}{Tomer Ullman}, \bibinfo{person}{Antonio Torralba}, {and} \bibinfo{person}{Joshua~B Tenenbaum}.} \bibinfo{year}{2016}\natexlab{}.
\newblock \showarticletitle{A compositional object-based approach to learning physical dynamics}.
\newblock \bibinfo{journal}{\emph{arXiv preprint arXiv:1612.00341}} (\bibinfo{year}{2016}).
\newblock


\bibitem[Chen et~al\mbox{.}(2022)]%
        {chen2022imitation}
\bibfield{author}{\bibinfo{person}{Siwei Chen}, \bibinfo{person}{Xiao Ma}, {and} \bibinfo{person}{Zhongwen Xu}.} \bibinfo{year}{2022}\natexlab{}.
\newblock \bibinfo{title}{Imitation Learning via Differentiable Physics}.
\newblock
\newblock
\showeprint[arxiv]{2206.04873}~[cs.LG]


\bibitem[Chen et~al\mbox{.}(2023)]%
        {chen2023daxbench}
\bibfield{author}{\bibinfo{person}{Siwei Chen}, \bibinfo{person}{Yiqing Xu}, \bibinfo{person}{Cunjun Yu}, \bibinfo{person}{Linfeng Li}, \bibinfo{person}{Xiao Ma}, \bibinfo{person}{Zhongwen Xu}, {and} \bibinfo{person}{David Hsu}.} \bibinfo{year}{2023}\natexlab{}.
\newblock \showarticletitle{DaxBench: Benchmarking Deformable Object Manipulation with Differentiable Physics}. In \bibinfo{booktitle}{\emph{The Eleventh International Conference on Learning Representations}}.
\newblock
\urldef\tempurl%
\url{https://openreview.net/forum?id=1NAzMofMnWl}
\showURL{%
\tempurl}


\bibitem[Coumans and Bai(2016)]%
        {pybullet}
\bibfield{author}{\bibinfo{person}{Erwin Coumans} {and} \bibinfo{person}{Yunfei Bai}.} \bibinfo{year}{2016}\natexlab{}.
\newblock \showarticletitle{Pybullet, a python module for physics simulation for games, robotics and machine learning}.
\newblock  (\bibinfo{year}{2016}).
\newblock


\bibitem[de~Avila Belbute-Peres et~al\mbox{.}(2018)]%
        {e2ediffctrl}
\bibfield{author}{\bibinfo{person}{Filipe de Avila Belbute-Peres}, \bibinfo{person}{Kevin Smith}, \bibinfo{person}{Kelsey Allen}, \bibinfo{person}{Josh Tenenbaum}, {and} \bibinfo{person}{J~Zico Kolter}.} \bibinfo{year}{2018}\natexlab{}.
\newblock \showarticletitle{End-to-end differentiable physics for learning and control}.
\newblock \bibinfo{journal}{\emph{Advances in neural information processing systems}}  \bibinfo{volume}{31} (\bibinfo{year}{2018}).
\newblock


\bibitem[Degrave et~al\mbox{.}(2019)]%
        {diffcontroller}
\bibfield{author}{\bibinfo{person}{Jonas Degrave}, \bibinfo{person}{Michiel Hermans}, \bibinfo{person}{Joni Dambre}, {et~al\mbox{.}}} \bibinfo{year}{2019}\natexlab{}.
\newblock \showarticletitle{A differentiable physics engine for deep learning in robotics}.
\newblock \bibinfo{journal}{\emph{Frontiers in neurorobotics}} (\bibinfo{year}{2019}), \bibinfo{pages}{6}.
\newblock


\bibitem[Erez et~al\mbox{.}(2015)]%
        {comparesim}
\bibfield{author}{\bibinfo{person}{Tom Erez}, \bibinfo{person}{Yuval Tassa}, {and} \bibinfo{person}{Emanuel Todorov}.} \bibinfo{year}{2015}\natexlab{}.
\newblock \showarticletitle{Simulation tools for model-based robotics: Comparison of bullet, havok, mujoco, ode and physx}. In \bibinfo{booktitle}{\emph{2015 IEEE international conference on robotics and automation (ICRA)}}. IEEE, \bibinfo{pages}{4397--4404}.
\newblock


\bibitem[Freeman et~al\mbox{.}(2021)]%
        {brax}
\bibfield{author}{\bibinfo{person}{C.~Daniel Freeman}, \bibinfo{person}{Erik Frey}, \bibinfo{person}{Anton Raichuk}, \bibinfo{person}{Sertan Girgin}, \bibinfo{person}{Igor Mordatch}, {and} \bibinfo{person}{Olivier Bachem}.} \bibinfo{year}{2021}\natexlab{}.
\newblock \showarticletitle{Brax - A Differentiable Physics Engine for Large Scale Rigid Body Simulation}.
\newblock \bibinfo{journal}{\emph{ArXiv}}  \bibinfo{volume}{abs/2106.13281} (\bibinfo{year}{2021}).
\newblock


\bibitem[Head et~al\mbox{.}(2018)]%
        {skopt}
\bibfield{author}{\bibinfo{person}{Tim Head}, \bibinfo{person}{MechCoder}, \bibinfo{person}{Gilles Louppe}, \bibinfo{person}{Iaroslav Shcherbatyi}, \bibinfo{person}{fcharras}, \bibinfo{person}{Zé Vinícius}, \bibinfo{person}{cmmalone}, \bibinfo{person}{Christopher Schröder}, \bibinfo{person}{nel215}, \bibinfo{person}{Nuno Campos}, \bibinfo{person}{Todd Young}, \bibinfo{person}{Stefano Cereda}, \bibinfo{person}{Thomas Fan}, \bibinfo{person}{rene rex}, \bibinfo{person}{Kejia~(KJ) Shi}, \bibinfo{person}{Justus Schwabedal}, \bibinfo{person}{carlosdanielcsantos}, \bibinfo{person}{Hvass-Labs}, \bibinfo{person}{Mikhail Pak}, \bibinfo{person}{SoManyUsernamesTaken}, \bibinfo{person}{Fred Callaway}, \bibinfo{person}{Loïc Estève}, \bibinfo{person}{Lilian Besson}, \bibinfo{person}{Mehdi Cherti}, \bibinfo{person}{Karlson Pfannschmidt}, \bibinfo{person}{Fabian Linzberger}, \bibinfo{person}{Christophe Cauet}, \bibinfo{person}{Anna Gut}, \bibinfo{person}{Andreas Mueller}, {and} \bibinfo{person}{Alexander Fabisch}.}
  \bibinfo{year}{2018}\natexlab{}.
\newblock \bibinfo{title}{scikit-optimize/scikit-optimize: v0.5.2}.
\newblock
\newblock
\urldef\tempurl%
\url{https://doi.org/10.5281/zenodo.1207017}
\showDOI{\tempurl}


\bibitem[Heiden et~al\mbox{.}(2021)]%
        {disect}
\bibfield{author}{\bibinfo{person}{Eric Heiden}, \bibinfo{person}{Miles Macklin}, \bibinfo{person}{Yashraj~S. Narang}, \bibinfo{person}{Dieter Fox}, \bibinfo{person}{Animesh Garg}, {and} \bibinfo{person}{Fabio Ramos}.} \bibinfo{year}{2021}\natexlab{}.
\newblock \showarticletitle{DiSECt: A Differentiable Simulation Engine for Autonomous Robotic Cutting}.
\newblock \bibinfo{journal}{\emph{ArXiv}}  \bibinfo{volume}{abs/2105.12244} (\bibinfo{year}{2021}).
\newblock
\urldef\tempurl%
\url{https://api.semanticscholar.org/CorpusID:235195762}
\showURL{%
\tempurl}


\bibitem[Hu et~al\mbox{.}(2018)]%
        {chainqueen}
\bibfield{author}{\bibinfo{person}{Yuanming Hu}, \bibinfo{person}{Jiancheng Liu}, \bibinfo{person}{Andrew~Everett Spielberg}, \bibinfo{person}{Joshua~B. Tenenbaum}, \bibinfo{person}{William~T. Freeman}, \bibinfo{person}{Jiajun Wu}, \bibinfo{person}{Daniela Rus}, {and} \bibinfo{person}{Wojciech Matusik}.} \bibinfo{year}{2018}\natexlab{}.
\newblock \showarticletitle{ChainQueen: A Real-Time Differentiable Physical Simulator for Soft Robotics}.
\newblock \bibinfo{journal}{\emph{2019 International Conference on Robotics and Automation (ICRA)}} (\bibinfo{year}{2018}), \bibinfo{pages}{6265--6271}.
\newblock
\urldef\tempurl%
\url{https://api.semanticscholar.org/CorpusID:52911940}
\showURL{%
\tempurl}


\bibitem[Ioffe and Szegedy(2015)]%
        {bn}
\bibfield{author}{\bibinfo{person}{Sergey Ioffe} {and} \bibinfo{person}{Christian Szegedy}.} \bibinfo{year}{2015}\natexlab{}.
\newblock \showarticletitle{Batch Normalization: Accelerating Deep Network Training by Reducing Internal Covariate Shift}.
\newblock \bibinfo{journal}{\emph{ArXiv}}  \bibinfo{volume}{abs/1502.03167} (\bibinfo{year}{2015}).
\newblock


\bibitem[Isenberg(2023)]%
        {Isenberg2023SystemIO}
\bibfield{author}{\bibinfo{person}{Douglas~R. Isenberg}.} \bibinfo{year}{2023}\natexlab{}.
\newblock \showarticletitle{System Identification of a Mobile Robot with Motion Capture Data}.
\newblock \bibinfo{journal}{\emph{2023 Intermountain Engineering, Technology and Computing (IETC)}} (\bibinfo{year}{2023}), \bibinfo{pages}{132--137}.
\newblock
\urldef\tempurl%
\url{https://api.semanticscholar.org/CorpusID:259215958}
\showURL{%
\tempurl}


\bibitem[Jiang and Liu(2018)]%
        {contactmodel}
\bibfield{author}{\bibinfo{person}{Yifeng Jiang} {and} \bibinfo{person}{C.~Karen Liu}.} \bibinfo{year}{2018}\natexlab{}.
\newblock \showarticletitle{Data-Augmented Contact Model for Rigid Body Simulation}. In \bibinfo{booktitle}{\emph{Conference on Learning for Dynamics \& Control}}.
\newblock
\urldef\tempurl%
\url{https://api.semanticscholar.org/CorpusID:3881096}
\showURL{%
\tempurl}


\bibitem[Khan et~al\mbox{.}(2022)]%
        {khan2022dynamic}
\bibfield{author}{\bibinfo{person}{Muhammad~Aseer Khan}, \bibinfo{person}{Dur-e-Zehra Baig}, \bibinfo{person}{Bilal Ashraf}, \bibinfo{person}{Husan Ali}, \bibinfo{person}{Junaid Rashid}, {and} \bibinfo{person}{Jungeun Kim}.} \bibinfo{year}{2022}\natexlab{}.
\newblock \showarticletitle{Dynamic modeling of a nonlinear two-wheeled robot using data-driven approach}.
\newblock \bibinfo{journal}{\emph{Processes}} \bibinfo{volume}{10}, \bibinfo{number}{3} (\bibinfo{year}{2022}), \bibinfo{pages}{524}.
\newblock


\bibitem[Koenig and Howard(2004)]%
        {gazebo}
\bibfield{author}{\bibinfo{person}{Nathan~P. Koenig} {and} \bibinfo{person}{Andrew Howard}.} \bibinfo{year}{2004}\natexlab{}.
\newblock \showarticletitle{Design and use paradigms for Gazebo, an open-source multi-robot simulator}.
\newblock \bibinfo{journal}{\emph{2004 IEEE/RSJ International Conference on Intelligent Robots and Systems (IROS) (IEEE Cat. No.04CH37566)}}  \bibinfo{volume}{3} (\bibinfo{year}{2004}), \bibinfo{pages}{2149--2154}.
\newblock
\urldef\tempurl%
\url{https://api.semanticscholar.org/CorpusID:206941306}
\showURL{%
\tempurl}


\bibitem[Kosmann-Schwarzbach et~al\mbox{.}(2011)]%
        {noether}
\bibfield{author}{\bibinfo{person}{Yvette Kosmann-Schwarzbach}, \bibinfo{person}{Bertram~E Schwarzbach}, {and} \bibinfo{person}{Yvette Kosmann-Schwarzbach}.} \bibinfo{year}{2011}\natexlab{}.
\newblock \bibinfo{booktitle}{\emph{The Noether Theorems}}.
\newblock \bibinfo{publisher}{Springer}.
\newblock


\bibitem[Lai et~al\mbox{.}(2020)]%
        {Lai2020BidirectionalMP}
\bibfield{author}{\bibinfo{person}{Hang Lai}, \bibinfo{person}{Jian Shen}, \bibinfo{person}{Weinan Zhang}, {and} \bibinfo{person}{Yong Yu}.} \bibinfo{year}{2020}\natexlab{}.
\newblock \showarticletitle{Bidirectional Model-based Policy Optimization}.
\newblock \bibinfo{journal}{\emph{ArXiv}}  \bibinfo{volume}{abs/2007.01995} (\bibinfo{year}{2020}).
\newblock
\urldef\tempurl%
\url{https://api.semanticscholar.org/CorpusID:220364497}
\showURL{%
\tempurl}


\bibitem[Li et~al\mbox{.}(2021)]%
        {diffcloth}
\bibfield{author}{\bibinfo{person}{Yifei Li}, \bibinfo{person}{Tao Du}, \bibinfo{person}{Kui Wu}, \bibinfo{person}{Jie Xu}, {and} \bibinfo{person}{Wojciech Matusik}.} \bibinfo{year}{2021}\natexlab{}.
\newblock \showarticletitle{DiffCloth: Differentiable Cloth Simulation with Dry Frictional Contact}.
\newblock \bibinfo{journal}{\emph{ACM Transactions on Graphics (TOG)}}  \bibinfo{volume}{42} (\bibinfo{year}{2021}), \bibinfo{pages}{1 -- 20}.
\newblock
\urldef\tempurl%
\url{https://api.semanticscholar.org/CorpusID:235390650}
\showURL{%
\tempurl}


\bibitem[Li et~al\mbox{.}(2018)]%
        {particledyn}
\bibfield{author}{\bibinfo{person}{Yunzhu Li}, \bibinfo{person}{Jiajun Wu}, \bibinfo{person}{Russ Tedrake}, \bibinfo{person}{Joshua~B. Tenenbaum}, {and} \bibinfo{person}{Antonio Torralba}.} \bibinfo{year}{2018}\natexlab{}.
\newblock \showarticletitle{Learning Particle Dynamics for Manipulating Rigid Bodies, Deformable Objects, and Fluids}.
\newblock \bibinfo{journal}{\emph{ArXiv}}  \bibinfo{volume}{abs/1810.01566} (\bibinfo{year}{2018}).
\newblock
\urldef\tempurl%
\url{https://api.semanticscholar.org/CorpusID:52917627}
\showURL{%
\tempurl}


\bibitem[Lin et~al\mbox{.}(2021)]%
        {modnas}
\bibfield{author}{\bibinfo{person}{Yunfeng Lin}, \bibinfo{person}{Guilin Li}, \bibinfo{person}{Xingzhi Zhang}, \bibinfo{person}{Weinan Zhang}, \bibinfo{person}{Bo Chen}, \bibinfo{person}{Bo Chen}, \bibinfo{person}{Ruiming Tang}, \bibinfo{person}{Zhenguo Li}, \bibinfo{person}{Jiashi Feng}, {and} \bibinfo{person}{Yong Yu}.} \bibinfo{year}{2021}\natexlab{}.
\newblock \showarticletitle{ModularNAS: Towards Modularized and Reusable Neural Architecture Search}. In \bibinfo{booktitle}{\emph{Conference on Machine Learning and Systems}}.
\newblock


\bibitem[Makoviychuk et~al\mbox{.}(2021)]%
        {isaacgym}
\bibfield{author}{\bibinfo{person}{Viktor Makoviychuk}, \bibinfo{person}{Lukasz Wawrzyniak}, \bibinfo{person}{Yunrong Guo}, \bibinfo{person}{Michelle Lu}, \bibinfo{person}{Kier Storey}, \bibinfo{person}{Miles Macklin}, \bibinfo{person}{David Hoeller}, \bibinfo{person}{N. Rudin}, \bibinfo{person}{Arthur Allshire}, \bibinfo{person}{Ankur Handa}, {and} \bibinfo{person}{Gavriel State}.} \bibinfo{year}{2021}\natexlab{}.
\newblock \showarticletitle{Isaac Gym: High Performance GPU-Based Physics Simulation For Robot Learning}.
\newblock \bibinfo{journal}{\emph{ArXiv}}  \bibinfo{volume}{abs/2108.10470} (\bibinfo{year}{2021}).
\newblock


\bibitem[Mittal et~al\mbox{.}(2023)]%
        {orbit}
\bibfield{author}{\bibinfo{person}{Mayank Mittal}, \bibinfo{person}{Calvin Yu}, \bibinfo{person}{Qinxi Yu}, \bibinfo{person}{Jingzhou Liu}, \bibinfo{person}{Nikita Rudin}, \bibinfo{person}{David Hoeller}, \bibinfo{person}{Jia~Lin Yuan}, \bibinfo{person}{Ritvik Singh}, \bibinfo{person}{Yunrong Guo}, \bibinfo{person}{Hammad Mazhar}, {et~al\mbox{.}}} \bibinfo{year}{2023}\natexlab{}.
\newblock \showarticletitle{Orbit: A unified simulation framework for interactive robot learning environments}.
\newblock \bibinfo{journal}{\emph{IEEE Robotics and Automation Letters}} (\bibinfo{year}{2023}).
\newblock


\bibitem[Mrowca et~al\mbox{.}(2018)]%
        {flexphys}
\bibfield{author}{\bibinfo{person}{Damian Mrowca}, \bibinfo{person}{Chengxu Zhuang}, \bibinfo{person}{Elias Wang}, \bibinfo{person}{Nick Haber}, \bibinfo{person}{Li~F Fei-Fei}, \bibinfo{person}{Josh Tenenbaum}, {and} \bibinfo{person}{Daniel~L Yamins}.} \bibinfo{year}{2018}\natexlab{}.
\newblock \showarticletitle{Flexible neural representation for physics prediction}.
\newblock \bibinfo{journal}{\emph{Advances in neural information processing systems}}  \bibinfo{volume}{31} (\bibinfo{year}{2018}).
\newblock


\bibitem[Murthy et~al\mbox{.}(2020)]%
        {gradsim}
\bibfield{author}{\bibinfo{person}{J~Krishna Murthy}, \bibinfo{person}{Miles Macklin}, \bibinfo{person}{Florian Golemo}, \bibinfo{person}{Vikram Voleti}, \bibinfo{person}{Linda Petrini}, \bibinfo{person}{Martin Weiss}, \bibinfo{person}{Breandan Considine}, \bibinfo{person}{J{\'e}r{\^o}me Parent-L{\'e}vesque}, \bibinfo{person}{Kevin Xie}, \bibinfo{person}{Kenny Erleben}, {et~al\mbox{.}}} \bibinfo{year}{2020}\natexlab{}.
\newblock \showarticletitle{gradsim: Differentiable simulation for system identification and visuomotor control}. In \bibinfo{booktitle}{\emph{International conference on learning representations}}.
\newblock


\bibitem[Mylvaganam and Sassano(2018)]%
        {multiwmr}
\bibfield{author}{\bibinfo{person}{Thulasi Mylvaganam} {and} \bibinfo{person}{Mario Sassano}.} \bibinfo{year}{2018}\natexlab{}.
\newblock \showarticletitle{Autonomous collision avoidance for wheeled mobile robots using a differential game approach}.
\newblock \bibinfo{journal}{\emph{European Journal of Control}}  \bibinfo{volume}{40} (\bibinfo{year}{2018}), \bibinfo{pages}{53--61}.
\newblock


\bibitem[Nvidia({[n.\,d.]})]%
        {physx}
\bibfield{author}{\bibinfo{person}{Nvidia}.} \bibinfo{year}{[n.\,d.]}\natexlab{}.
\newblock \bibinfo{booktitle}{\emph{Nvidia PhysX}}.
\newblock
\urldef\tempurl%
\url{https://developer.nvidia.com/physx-sdk}
\showURL{%
\tempurl}


\bibitem[Paszke et~al\mbox{.}(2019)]%
        {pytorch}
\bibfield{author}{\bibinfo{person}{Adam Paszke}, \bibinfo{person}{Sam Gross}, \bibinfo{person}{Francisco Massa}, \bibinfo{person}{Adam Lerer}, \bibinfo{person}{James Bradbury}, \bibinfo{person}{Gregory Chanan}, \bibinfo{person}{Trevor Killeen}, \bibinfo{person}{Zeming Lin}, \bibinfo{person}{Natalia Gimelshein}, \bibinfo{person}{Luca Antiga}, \bibinfo{person}{Alban Desmaison}, \bibinfo{person}{Andreas K{\"o}pf}, \bibinfo{person}{Edward Yang}, \bibinfo{person}{Zach DeVito}, \bibinfo{person}{Martin Raison}, \bibinfo{person}{Alykhan Tejani}, \bibinfo{person}{Sasank Chilamkurthy}, \bibinfo{person}{Benoit Steiner}, \bibinfo{person}{Lu Fang}, \bibinfo{person}{Junjie Bai}, {and} \bibinfo{person}{Soumith Chintala}.} \bibinfo{year}{2019}\natexlab{}.
\newblock \showarticletitle{PyTorch: An Imperative Style, High-Performance Deep Learning Library}. In \bibinfo{booktitle}{\emph{NeurIPS}}.
\newblock


\bibitem[Quigley et~al\mbox{.}(2009)]%
        {ros}
\bibfield{author}{\bibinfo{person}{Morgan Quigley}, \bibinfo{person}{Ken Conley}, \bibinfo{person}{Brian Gerkey}, \bibinfo{person}{Josh Faust}, \bibinfo{person}{Tully Foote}, \bibinfo{person}{Jeremy Leibs}, \bibinfo{person}{Rob Wheeler}, \bibinfo{person}{Andrew~Y Ng}, {et~al\mbox{.}}} \bibinfo{year}{2009}\natexlab{}.
\newblock \showarticletitle{ROS: an open-source Robot Operating System}. In \bibinfo{booktitle}{\emph{ICRA workshop on open source software}}, Vol.~\bibinfo{volume}{3}. Kobe, Japan, \bibinfo{pages}{5}.
\newblock


\bibitem[Rudin et~al\mbox{.}(2021)]%
        {walkinmin}
\bibfield{author}{\bibinfo{person}{N. Rudin}, \bibinfo{person}{David Hoeller}, \bibinfo{person}{Philipp Reist}, {and} \bibinfo{person}{Marco Hutter}.} \bibinfo{year}{2021}\natexlab{}.
\newblock \showarticletitle{Learning to Walk in Minutes Using Massively Parallel Deep Reinforcement Learning}.
\newblock \bibinfo{journal}{\emph{ArXiv}}  \bibinfo{volume}{abs/2109.11978} (\bibinfo{year}{2021}).
\newblock
\urldef\tempurl%
\url{https://api.semanticscholar.org/CorpusID:237635100}
\showURL{%
\tempurl}


\bibitem[Sanchez-Gonzalez et~al\mbox{.}(2020)]%
        {gnnphys}
\bibfield{author}{\bibinfo{person}{Alvaro Sanchez-Gonzalez}, \bibinfo{person}{Jonathan Godwin}, \bibinfo{person}{Tobias Pfaff}, \bibinfo{person}{Rex Ying}, \bibinfo{person}{Jure Leskovec}, {and} \bibinfo{person}{Peter~W. Battaglia}.} \bibinfo{year}{2020}\natexlab{}.
\newblock \showarticletitle{Learning to Simulate Complex Physics with Graph Networks}.
\newblock \bibinfo{journal}{\emph{ArXiv}}  \bibinfo{volume}{abs/2002.09405} (\bibinfo{year}{2020}).
\newblock
\urldef\tempurl%
\url{https://api.semanticscholar.org/CorpusID:211252550}
\showURL{%
\tempurl}


\bibitem[Sherstinsky(2020)]%
        {rnn}
\bibfield{author}{\bibinfo{person}{Alex Sherstinsky}.} \bibinfo{year}{2020}\natexlab{}.
\newblock \showarticletitle{Fundamentals of recurrent neural network (RNN) and long short-term memory (LSTM) network}.
\newblock \bibinfo{journal}{\emph{Physica D: Nonlinear Phenomena}}  \bibinfo{volume}{404} (\bibinfo{year}{2020}), \bibinfo{pages}{132306}.
\newblock


\bibitem[Shi et~al\mbox{.}(2015)]%
        {lstm}
\bibfield{author}{\bibinfo{person}{Xingjian Shi}, \bibinfo{person}{Zhourong Chen}, \bibinfo{person}{Hao Wang}, \bibinfo{person}{D.~Y. Yeung}, \bibinfo{person}{Wai-Kin Wong}, {and} \bibinfo{person}{Wang chun Woo}.} \bibinfo{year}{2015}\natexlab{}.
\newblock \showarticletitle{Convolutional LSTM Network: A Machine Learning Approach for Precipitation Nowcasting}. In \bibinfo{booktitle}{\emph{NIPS}}.
\newblock


\bibitem[Siciliano et~al\mbox{.}(2008)]%
        {robothandbook}
\bibfield{author}{\bibinfo{person}{Bruno Siciliano}, \bibinfo{person}{Oussama Khatib}, {and} \bibinfo{person}{Torsten Kr{\"o}ger}.} \bibinfo{year}{2008}\natexlab{}.
\newblock \bibinfo{booktitle}{\emph{Springer handbook of robotics}}. Vol.~\bibinfo{volume}{200}.
\newblock \bibinfo{publisher}{Springer}.
\newblock


\bibitem[Smith et~al\mbox{.}(2005)]%
        {ode}
\bibfield{author}{\bibinfo{person}{Russell Smith} {et~al\mbox{.}}} \bibinfo{year}{2005}\natexlab{}.
\newblock \showarticletitle{Open dynamics engine}.
\newblock  (\bibinfo{year}{2005}).
\newblock


\bibitem[Sturm et~al\mbox{.}(2012)]%
        {Sturm2012ABF}
\bibfield{author}{\bibinfo{person}{J{\"u}rgen Sturm}, \bibinfo{person}{Nikolas Engelhard}, \bibinfo{person}{Felix Endres}, \bibinfo{person}{Wolfram Burgard}, {and} \bibinfo{person}{Daniel Cremers}.} \bibinfo{year}{2012}\natexlab{}.
\newblock \showarticletitle{A benchmark for the evaluation of RGB-D SLAM systems}.
\newblock \bibinfo{journal}{\emph{2012 IEEE/RSJ International Conference on Intelligent Robots and Systems}} (\bibinfo{year}{2012}), \bibinfo{pages}{573--580}.
\newblock
\urldef\tempurl%
\url{https://api.semanticscholar.org/CorpusID:206942855}
\showURL{%
\tempurl}


\bibitem[Todorov et~al\mbox{.}(2012)]%
        {mujoco}
\bibfield{author}{\bibinfo{person}{Emanuel Todorov}, \bibinfo{person}{Tom Erez}, {and} \bibinfo{person}{Yuval Tassa}.} \bibinfo{year}{2012}\natexlab{}.
\newblock \showarticletitle{MuJoCo: A physics engine for model-based control}.
\newblock \bibinfo{journal}{\emph{2012 IEEE/RSJ International Conference on Intelligent Robots and Systems}} (\bibinfo{year}{2012}), \bibinfo{pages}{5026--5033}.
\newblock


\bibitem[Uddin(2020)]%
        {Uddin2020SystemIO}
\bibfield{author}{\bibinfo{person}{Nur Uddin}.} \bibinfo{year}{2020}\natexlab{}.
\newblock \showarticletitle{System Identification of Two-Wheeled Robot Dynamics Using Neural Networks}.
\newblock \bibinfo{journal}{\emph{Journal of Physics: Conference Series}}  \bibinfo{volume}{1577} (\bibinfo{year}{2020}).
\newblock
\urldef\tempurl%
\url{https://api.semanticscholar.org/CorpusID:225567177}
\showURL{%
\tempurl}


\bibitem[Wen et~al\mbox{.}(2023)]%
        {Wen2023LargeSM}
\bibfield{author}{\bibinfo{person}{Muning Wen}, \bibinfo{person}{Runji Lin}, \bibinfo{person}{Hanjing Wang}, \bibinfo{person}{Yaodong Yang}, \bibinfo{person}{Ying Wen}, \bibinfo{person}{Luo Mai}, \bibinfo{person}{J. Wang}, \bibinfo{person}{Haifeng Zhang}, {and} \bibinfo{person}{Weinan Zhang}.} \bibinfo{year}{2023}\natexlab{}.
\newblock \showarticletitle{Large sequence models for sequential decision-making: a survey}.
\newblock \bibinfo{journal}{\emph{Frontiers of Computer Science}}  \bibinfo{volume}{17} (\bibinfo{year}{2023}).
\newblock
\urldef\tempurl%
\url{https://api.semanticscholar.org/CorpusID:259252117}
\showURL{%
\tempurl}


\bibitem[Wu et~al\mbox{.}(2019)]%
        {Wu2019ModelIF}
\bibfield{author}{\bibinfo{person}{Yueh-Hua Wu}, \bibinfo{person}{Ting-Han Fan}, \bibinfo{person}{Peter~J. Ramadge}, {and} \bibinfo{person}{Hao Su}.} \bibinfo{year}{2019}\natexlab{}.
\newblock \showarticletitle{Model Imitation for Model-Based Reinforcement Learning}.
\newblock \bibinfo{journal}{\emph{ArXiv}}  \bibinfo{volume}{abs/1909.11821} (\bibinfo{year}{2019}).
\newblock
\urldef\tempurl%
\url{https://api.semanticscholar.org/CorpusID:202889287}
\showURL{%
\tempurl}


\bibitem[Xu et~al\mbox{.}(2021)]%
        {shac}
\bibfield{author}{\bibinfo{person}{Jie Xu}, \bibinfo{person}{Viktor Makoviychuk}, \bibinfo{person}{Yashraj Narang}, \bibinfo{person}{Fabio Ramos}, \bibinfo{person}{Wojciech Matusik}, \bibinfo{person}{Animesh Garg}, {and} \bibinfo{person}{Miles Macklin}.} \bibinfo{year}{2021}\natexlab{}.
\newblock \showarticletitle{Accelerated Policy Learning with Parallel Differentiable Simulation}. In \bibinfo{booktitle}{\emph{International Conference on Learning Representations}}.
\newblock


\bibitem[Zamora et~al\mbox{.}(2021)]%
        {Zamora2021PODSPO}
\bibfield{author}{\bibinfo{person}{Miguel Zamora}, \bibinfo{person}{Momchil Peychev}, \bibinfo{person}{Sehoon Ha}, \bibinfo{person}{Martin~T. Vechev}, {and} \bibinfo{person}{Stelian Coros}.} \bibinfo{year}{2021}\natexlab{}.
\newblock \showarticletitle{PODS: Policy Optimization via Differentiable Simulation}. In \bibinfo{booktitle}{\emph{International Conference on Machine Learning}}.
\newblock
\urldef\tempurl%
\url{https://api.semanticscholar.org/CorpusID:235826467}
\showURL{%
\tempurl}


\end{thebibliography}

\appendix

\newpage
\section{Dataset Visualizations}
\label{apdx:dset_plot}

\begin{figure}[htb]
\centering
    \includegraphics[width=.45\textwidth]{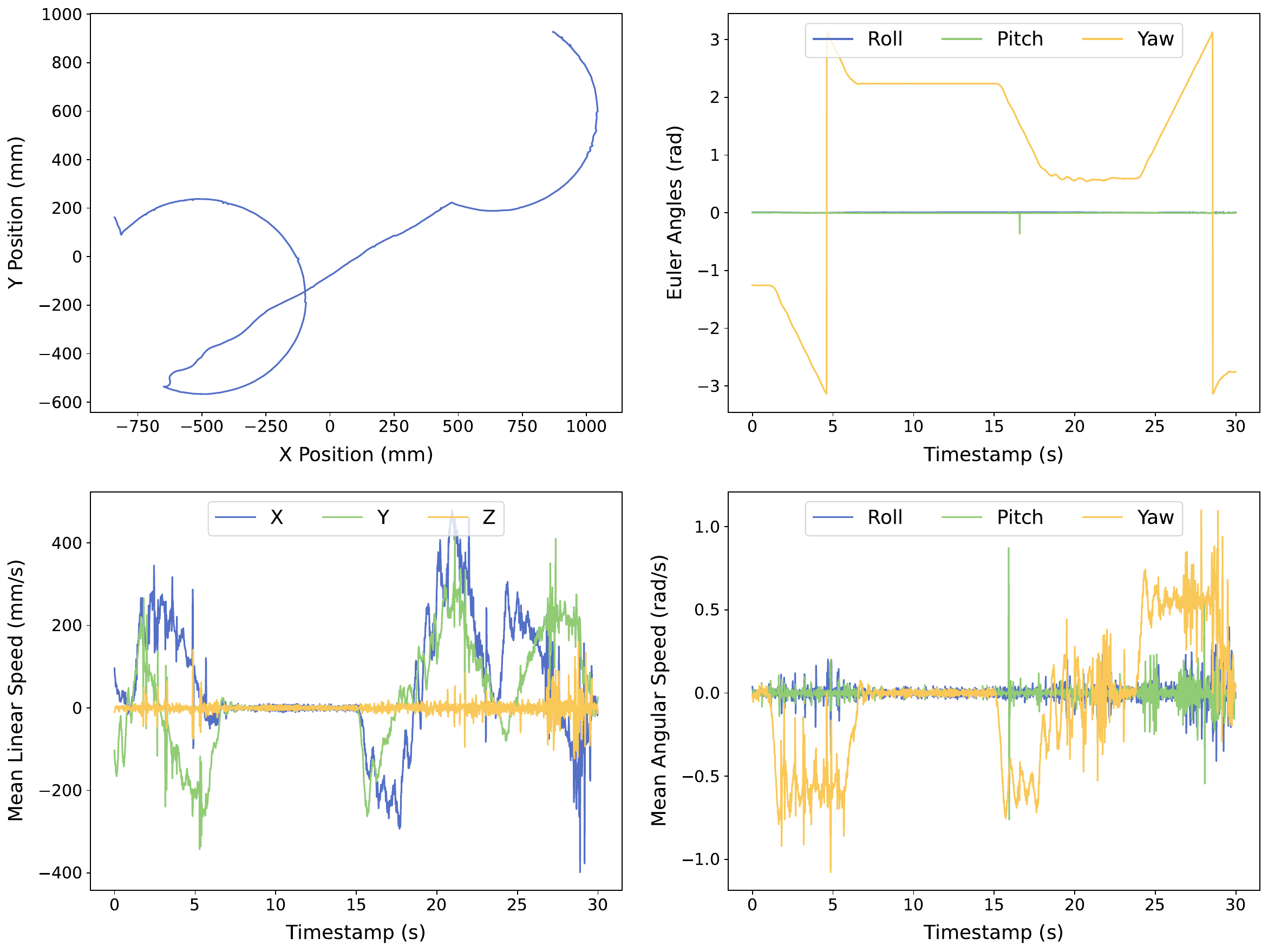}
    \caption{
Part of the raw pose data recorded from the robot actuated with random navigation commands.
The subplots (from left to right, top to bottom) are 1) X-Y positions; 2) Euler angles; 3) computed mean linear speed; 4) computed mean angular speed.
}
    \Description{.}
    \label{fig:_eval_dset}
\end{figure}

\begin{figure}[htb]
\centering
    \includegraphics[width=.45\textwidth]{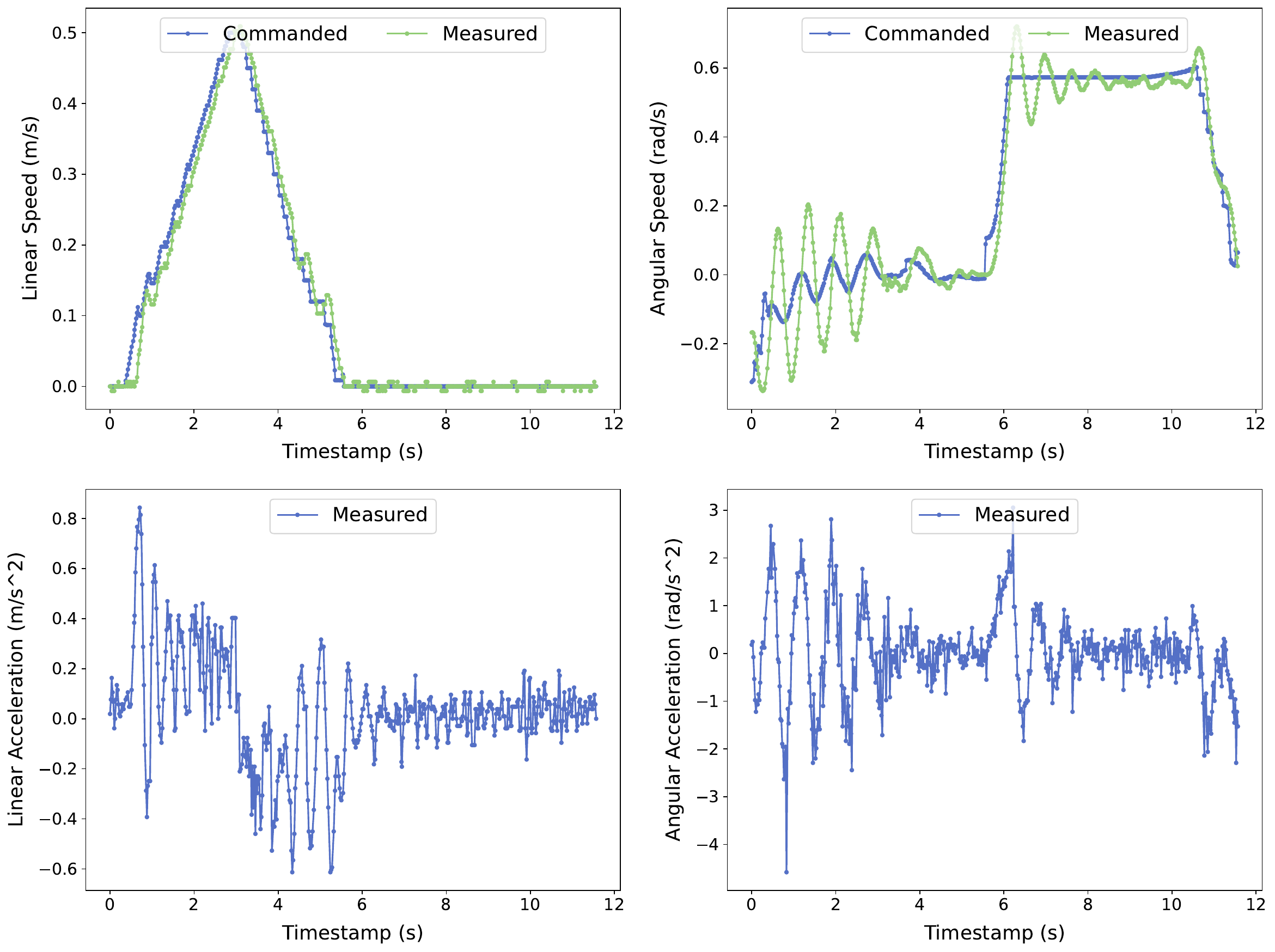}
    \caption{
Part of the robot command and proprioceptive sensor data.
The subplots (from left to right, top to bottom) are 1) commanded and measured linear speed; 2) commanded and measured angular velocity; 3) measured linear acceleration; 4) measured angular acceleration.
}
    \Description{.}
    \label{fig:_eval_cmds}
\end{figure}

In Figure~\ref{fig:_eval_dset} and Figure~\ref{fig:_eval_cmds}, we provide extra visualizations of the robot pose data from the motion capture system, as well as the command and sensor data recorded within the robot itself.

\section{Simulation Details}

\begin{table}[ht]
    \caption{Isaac Sim parameters}
    \label{table:_apdx_sim_params}
    \begin{tabular}{p{0.2\textwidth}p{0.2\textwidth}}
      \toprule
      Name & Value \\
      \midrule
      Static friction & 0.5\\
      Dynamics friction & 4.0\\
      Restitution & 0.0\\
      Friction combine mode & max\\
        
      Bounce threshold & 2.0\\
      Enable CCD & True\\
      \bottomrule
\end{tabular}
\end{table}

\begin{figure}[htb]
\centering
    \includegraphics[width=.43\textwidth]{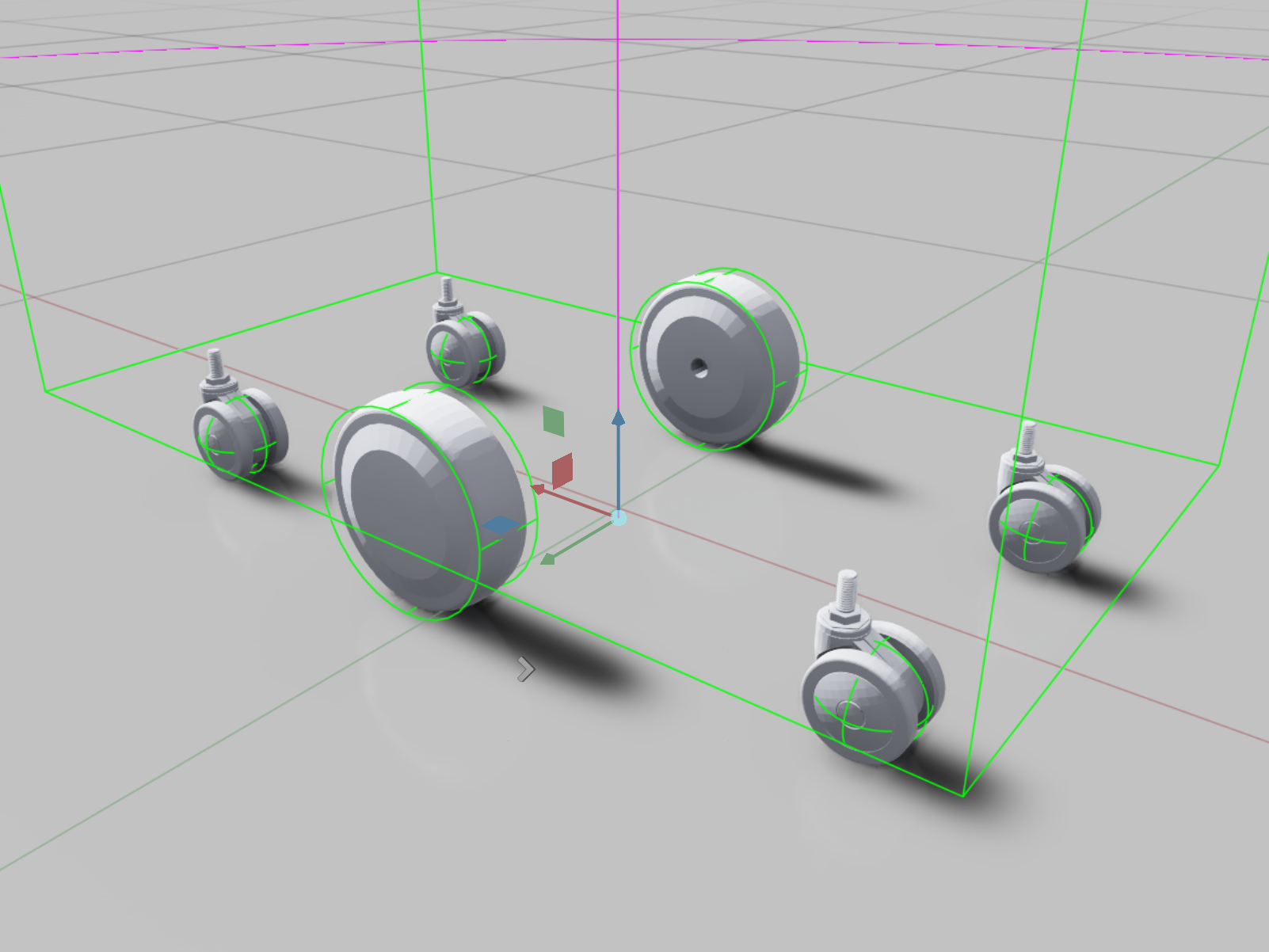}
    \caption{Rendering of the robot wheels in Isaac Sim simulations. The green wireframes show the shapes used as collision geometries for the chassis and the wheels.}
    \Description{.}
    \label{fig:_sim_mesh}
\end{figure}

The analytical simulation baseline is built on Isaac Sim 2022.1 and runs on NVIDIA RTX A2000 GPU.
We adopt the physical scene parameters in Table~\ref{table:_apdx_sim_params} to improve numerical stability and reduce prediction errors.
Physical bodies are simplified to provide extra stability and uniformity in simulations, see~Figure~\ref{fig:_sim_mesh}.

\section{Algorithm Details}

\subsection{Model Architectures}

We use a standard MLP architecture with ReLU activation and three hidden layers of size (32, 16, 8), respectively.
Batch normalization~\cite{bn} is enabled in training and evaluations.
We use a single linear layer with a bias parameter for Linear Regression models.

\subsection{Training Hyperparameters}

\begin{table}[htb]
    \caption{Training hyperparameters}
    \label{table:_apdx_hyper_params}
    \begin{tabular}{p{0.2\textwidth}p{0.2\textwidth}}
      \toprule
      Name & Value \\
      \midrule
      Initial LR & $5e^{-4}$ \\
      LR scheduler & Exponential\\
      LR gamma & 0.99999 \\
      Batch size & 200 \\
      L2 normalization & $1e^{-4}$ \\
      Early stopping patience & 32 \\
      \bottomrule
\end{tabular}
\end{table}

We use the settings shown in Table~\ref{table:_apdx_hyper_params} to train the data-driven models.

\end{document}